\documentclass{article}

\usepackage{multirow}
\usepackage{tabularx}
\usepackage{amsmath}
\usepackage{subfigure}

\usepackage{stackengine}
\usepackage{graphicx}
\usepackage{ulem}
 \usepackage{booktabs}
 
\usepackage[margin=0.7in]{geometry}
\begin{document}

\title{Monitoring Simulated Physical Weakness Using Detailed Behavioral Features and Personalized Modeling}

\date{}
\maketitle

Longfei Chen (longfei.chen@ed.ac.uk)

Muhammad Ahmed Raza (m.a.raza@ed.ac.uk)

Craig Innes (craig.innes@ed.ac.uk)

Subramanian Ramamoorthy (s.ramamoorthy@ed.ac.uk)

Robert B. Fisher (rbf@inf.ed.ac.uk)

The University of Edinburgh, 10 Crichton Street, Edinburgh, Scotland, UK, EH8 9AB

\begin{abstract}
Aging and chronic conditions affect older adults’ daily lives, making the early detection of developing health issues crucial. Weakness, which is common across many conditions, can subtly alter physical movements and daily activities. However, these behavioral changes can be difficult to detect because they are gradual and often masked by natural day-to-day variability. {To isolate the behavioral phenotype of weakness while controlling for confounding factors, this study simulates physical weakness in healthy adults through exercise-induced fatigue, providing interpretable insights into potential behavioral indicators for long-term monitoring.} 
A non-intrusive camera sensor is used to monitor individuals' daily sitting and relaxing activities over multiple days, allowing us to observe behavioral changes before and after simulated weakness. 
The system captures fine-grained features related to body motion, inactivity, and environmental context in real time while prioritizing privacy. A Bayesian Network models the relationships among activities, contextual factors, and behavioral indicators. 
Fine-grained features, including non-dominant upper-body motion speed and scale, together with inactivity distribution, are most effective when used with a 300-second window. Personalized models achieve 0.97 accuracy at distinguishing simulated weak days from normal days, and no universal set of optimal features or activities is observed across participants.
\end{abstract}

\section{Introduction}
Natural aging and chronic health conditions significantly affect older adults’ daily lives \cite{o1, o2}. The morbidity associated with aging and the decline in physical and mental abilities are among the main causes of suffering in older adults \cite{o11}. Weakness (or asthenia) is a phenotype that accompanies many prevalent health conditions \cite{wk1, wk2, wk3}. It manifests as physical weakness or a lack of energy in specific body parts or the entire body. Such weakness can lead to changes in the way older adults move and perform daily tasks, as they may need to compensate for their physical limitations by using different muscles or adopting altered postures \cite{wk1}. These behavior changes provide valuable insights into the health status of older adults. Early detection of developing health conditions can be crucial for prompt intervention and treatment \cite{early}.
However, detecting behavioral changes related to weakness can be challenging, particularly because the signs of long-term progressive conditions are often subtle, especially in the early stages \cite{subtle1, hard1}. As a result, such changes may not be immediately noticeable through snapshot clinical assessments, or observations by caregivers, families, or even the individuals themselves \cite{s2, hard2, cv10}.

Sensors provide a reliable and objective means of assessing the condition of older adults. There is a growing emphasis on innovative smart medical devices as care transitioning from hospitals to homes. According to \cite{healthcost2}, the integration of AI-powered healthcare monitoring devices has the potential to reduce annual healthcare spending in the US by 5 to 10 percent. Health monitoring devices enhance a sense of safety and reduce isolation among older adults, while also saving caregivers time through automation \cite{older1}. These devices offer unobtrusive, long-term monitoring, outperforming traditional manual approaches \cite{t1}. 
Among the array of non-intrusive health monitoring devices available, cameras are deemed to be one of the most suitable options for long-term monitoring of the progression of conditions in older adults. Cameras provide rich information about individuals and their surroundings by capturing data in the form of videos or images. They can extract the kinematics of the subject at a clinically acceptable level \cite{clinical, clinical2}, provide functionality to detect the presence of humans, objects, or pets, and facilitate an analysis of the environmental context. These attributes make camera-based monitoring highly interpretable, since the extracted information can be used to provide semantic explanations \cite{t1}. For instance, if a person experiences a fall or loses consciousness (manifested by an absence of body movement), the camera can promptly convey a visual message of the post-event scenario to a designated individual to assess the situation and provide help. It is also important to note that camera sensors still pose challenges \cite{t1}, including privacy concerns, measurement accuracy, and adaptability to diverse environments, e.g., occlusions, low light.

Simulated health conditions are increasingly used in aging research as a controlled way to study behavioral markers associated with functional changes. Weakness in older adults often occurs alongside multimorbidity\footnote{Multimorbidity is the coexistence of two or more chronic conditions. A total of 67\% of older adults have multimorbidity, with the prevalence increasing with age: 50\% for those under 65, 62\% for those aged 65-74, and 81.5\% for those aged 85 and over.} \cite{fried2, wk5}, meaning that it is influenced by various confounding health factors.
To specifically focus on the behavioral phenotype of weakness while controlling for confounders, we simulate weakness in healthy adults through exercise-induced fatigue and observe their behavioral changes before and after the workout.
We believe that the weakness observed post-workout provides a reasonable approximation to the physical manifestations of natural weakness in older adults. 
Despite differing underlying causes, they share similar observable characteristics, such as reduced energy, slowed movement, decreased strength and endurance, altered movement patterns, reduced range of motion, increased resting or recovery time, and reduced physical activity \cite{fried2, wk5}. {The simulation allows us to isolate these behavioral features, while minimizing the influence of multimorbidity and other individual-specific conditions.}

\begin{figure}[htbp]
    \centering
    \includegraphics[width=.95\linewidth]{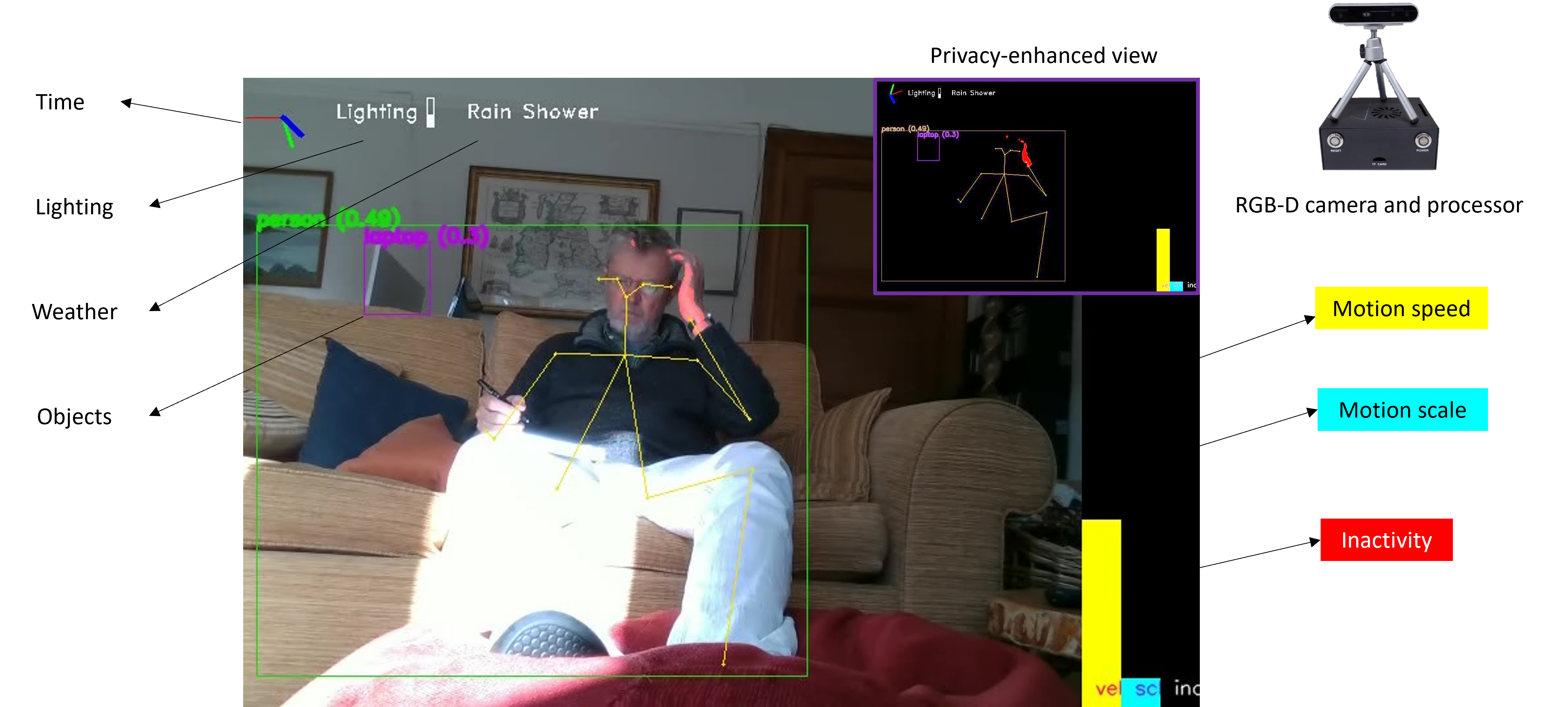}
    \caption{A compact system designed for monitoring older adults in their homes employs an RGB-D camera and a computer processor. This system prioritizes privacy by discarding image/video data after extracting the necessary information. Motion, inactivity features, and environmental context are extracted in real time. Detected movement pixels are shown in red.} 
    \label{fig:scenario}
\end{figure}

To capture behavioral data, we employ a single fixed RGB-D camera (see Fig. \ref{fig:scenario} for an example setup). Subjects are monitored while performing common daily activities within a designated area in a room. The system extracts body motion, inactivity, and environmental context features in real time. No visual frames are stored in order to preserve privacy. We model the dependencies between behavioral observations, activity types, and environmental context using a Bayesian Network. In addition, multiple classifiers and information-based methods are used to evaluate the importance of features and activities associated with simulated weakness. We also examine various temporal window sizes to determine the most effective time scale for observing behavioral changes and distinguishing between normal and weak states.

This study aims to address the following research questions:

1)	Can we accurately monitor changes in people’s behavior, specifically those caused by simulated weakness, during common daily activities?

2) Which behavioral features and activities exhibit the most pronounced changes?

3) What is the optimal temporal window for observing these changes?

{By investigating these questions, our objective is to understand whether automatically quantifying behavioral changes in a controlled, simulated scenario can provide insights into functional changes relevant to long-term health monitoring.}
Section \ref{sec:class} shows that participant-specific models can distinguish behavioral changes associated with simulated weakness.
Section \ref{sec:reA} demonstrates that certain features and activities are informative, although these vary across individuals.
Section \ref{sec:timescale} identifies the 300-second window as the most effective time scale for analysis.

\section{Related Works}  \label{sec:rw}
Changes in physical mobility can serve as valuable indicators for various common conditions in older adults \cite{decline}, including frailty, stroke, arthritis, Alzheimer's disease, depression, natural aging, and Parkinson's \cite{weakincon}. For example, frailty is characterized by weight loss, weakness, exhaustion, slowness, and low activity \cite{fried}. Stroke can lead to motor problems such as weakness, paralysis, and coordination and balance issues \cite{stroke}. Arthritis involves joint inflammation, leading to pain, stiffness, and reduced mobility, predominantly in the hands, knees, and hips. Multiple sclerosis may cause muscle weakness, stiffness, and tremors \cite{Arthscler}. Moreover, Alzheimer's disease can impact motor function, including coordination and balance difficulties \cite{Alz}. These conditions significantly influence the behaviors of older adults, ranging from mild to severe impairment, depending on disease severity and progression. As a result, older adults may encounter difficulties in performing daily activities that require strength and coordination. While many studies focus on detecting significant events, such as falls \cite{falls}, it is also crucial to recognize gradual health changes, especially considering the well-known phenomenon of change-blindness, where sufficiently slow changes are not perceptible by humans (but could be detected through automated record-keeping).

Sensors enable the creation of a comprehensive behavioral profile from continuous monitoring over the long-term \cite{s1}. These sensors encompass a variety of devices, including wearables \cite{wearsur1, wsur1}, smart home devices \cite{s8, f2}, and cameras \cite{cv1, cv10}. They have demonstrated their effectiveness in various healthcare applications. Some older adults view sensor-based monitoring as a highly beneficial approach that not only enhances their safety but also promotes independent living, enabling them to remain active while respecting their natural lifestyle choices \cite{older1, hard1}. In particular, "zero-interaction" sensing technologies are favored \cite{zeroi}.

\textbf{Sensors:}
Wearable sensors have emerged as valuable tools for measuring healthcare-related parameters in older adults. These sensors accurately measure human motion, localization, and tracking, making them suitable for various applications, such as frailty assessment, fall risk evaluation, monitoring chronic neurological diseases, promoting active living, and cognitive assessment~\cite{s1, wsur1}. A study demonstrated that inertial sensors effectively assess frailty, providing an objective measure of an individual's physical condition \cite{wearfra1}. Wearable inertial sensors have also been utilized to predict Functional Independence Measure scores in patients undergoing inpatient rehabilitation \cite{w1}. Moreover, they have been used to monitor turning movements associated with cognitive function in older adult participants \cite{w2}. However, the adoption of wearable sensor-based monitoring depends on both the ability of older adults with varying cognitive abilities to consistently wear and charge these devices, and the issues of stigma, which may impede widespread acceptance \cite{weardis1}.

Ambient sensors, often integrated into the living environment, encompass a range of devices such as passive infrared (PIR) motion sensors, magnet/contact switches, temperature sensors, light sensors, humidity sensors, vibration sensors, pressure sensors, and radio-frequency identification (RFID) sensors \cite{s1}. These sensors offer the capability to monitor diverse health-related parameters in older adults over extended periods. Smart home technology has been leveraged for healthcare applications, including assessing social isolation, cognitive health, functional health, and behavioral changes related to conditions such as radiation treatment, insomnia, depression, or dementia \cite{f3, f1, s5, s6, f2, s2}. However, smart homes face challenges due to the inherent complexity of the sensor system, in which each sensor typically serves a specific function, potentially leading to reliability and accuracy issues \cite{senserintro}. Monitoring multiple individuals simultaneously and distinguishing the impact of pets within the environment are also challenging tasks, primarily due to the limited ability to understand high-level semantic context \cite{homeissue}.

Camera sensors have emerged as powerful tools for monitoring the physical and contextual behavior of individuals, offering extensive and high-dimensional data capture capabilities. Recent advancements in computer vision algorithms and hardware have significantly enhanced their potential. In comparison to wearable sensors and basic ambient sensors, cameras provide a more comprehensive view of the monitored subject, offering rich, interpretable information about the individual, the surrounding environment, and their interactions \cite{cv1}. Notably, camera sensors are employed in various healthcare applications, including Parkinson's disease, stroke, epilepsy, and frailty assessments, as they can capture both upper and lower limb kinematic measurements \cite{cv1}. For example, researchers have utilized the Kinect RGB-D camera to monitor the body movements of pre-frail and frail elderly individuals \cite{cv3}. Wearable cameras have been employed to automatically identify sedentary periods in older adults, which have been associated with poor health outcomes \cite{cv2}. The fusion of camera data with other sensors, such as accelerometers, light and door sensors, microphones, and wearable devices, has been shown to effectively monitor dementia patients and detect long-term health changes in older adults \cite{DemCare, hard1}. Emerging impulse-radar sensors and depth sensors show promise for healthcare purposes as they do not require individuals to wear or operate any devices, offering similar information as video cameras while providing enhanced privacy protection \cite{g4, senserintro}.

\textbf{Sensor accuracy:}
Sensor selection plays a crucial role in smart home data analysis. 
Studies have shown that motion sensors are the most informative for activity recognition, with areas of high movement, such as the kitchen and living room, being particularly significant \cite{ss2}. Some research endeavors have employed multiple motion sensors with small fields of view to estimate walking speed when individuals pass by, alongside passive infrared motion sensors to detect changes in heat sources \cite{infre1}. However, ambient motion sensors may struggle to distinguish different levels of motion intensity, such as mild and high exertion of body parts \cite{infre1}. Wearable sensors exhibit high accuracy in estimating limb motion \cite{wearacc}, but they can be inconvenient for long-term use. Conversely, cameras can provide reasonably accurate motion estimates. For instance, a 3D markerless motion capture technique demonstrated that it could correctly reproduce the movements of participants within an accuracy of 30 mm \cite{clinical}. Another study compared upper-limb kinematics collected from a Kinect v1 against a 6-camera 3D marker-based system, where shoulder elevation angle had a reported difference of 3.32 degrees ± 2.80 degrees \cite{clinical2}. Both measurement errors were reported to be clinically acceptable. 

\textbf{Features:}
Features are numerical descriptions extracted from health monitoring data. They serve as indicators for identifying signs of disease, monitoring chronic conditions, and establishing personalized healthcare records. These features encompass various aspects of an individual's behavior and physiology.

Motor-related features encompass a wide range of parameters that shed light on an individual's physical capabilities. Metrics like walking distance and walking speed are key indicators of mobility and overall health \cite{s6, s9}. Stride length is an important measure related to walking patterns \cite{hard1}. Unconventional indicators such as patterns of using a computer mouse can provide insights into cognitive function \cite{s2}. Parameters such as speed, acceleration, and frequency of hand and wrist movements, trunk speed, and wrist speed offer valuable information \cite{cv1}. Studies have shown that slower execution speed may be detectable in the early stages of cognitive decline \cite{cv10}. For instance, in one study \cite{cv12}, participants' task execution time, walking speed, step length, and the number of errors or omissions in task completion were used to assess Alzheimer's disease symptoms.

Time-related features include measures such as activity density map across hours and days \cite{f1}, event/activity duration \cite{s5, s6, s9}, time of the day \cite{f2, cv2},  size of the sliding window \cite{f2}, time between two events \cite{f2}, amount of time spent outside the home \cite{f3, hard1, s6}, amount of time sitting \cite{s7, cv2}, the distribution of time spent in different home areas \cite{f3}. Studies have suggested that features like extended periods of sitting may be indicative of weakness \cite{f3}.

Activity-level features include information such as activity type \cite{cv2}, duration of activities \cite{f1}, number of sensor/event logs \cite{f1, s5}, number of complete tasks, interruptions, and omissions \cite{f1}, activity regularity \cite{s1}. Additionally, statistics such as minimum, maximum, sum, median, standard deviation, zero crossings, correlation, and skewness are often computed for these features \cite{s2, s9}. Leask et al. \cite{cv2} used activity type, environment, and interactions to understand sedentary behaviors in older adults. K\"onig et al. \cite{cv13} did feature selection on gait and event features extracted from a video event monitoring system for assessment of autonomy.

Lower-body features are widely employed for diverse purposes in older adult monitoring; however, they may not be suitable for some occasions such as when the older adults spend most of their time sitting at their favorite easy chair during the day \cite{cv2}. Moreover, the motion features extracted in the above-mentioned works are often quite coarse, and existing studies typically do not explore the relationships among these variables or model their dependencies, even though some variables may be highly correlated with others. 
In our study, we aim at daily sitting scenarios at home and extract fine-grained motion features of individuals, while modeling the relationship among behavioral features, the health states, and the environmental context.

\textbf{Activities:}
Activities also need to be carefully chosen to effectively monitor people's health states \cite{s5}. 
Some research uses specifically designed tasks to assess health states, which are clinically verified. In the study \cite{s5}, participants were asked to perform a sequence of 8 activities representing instrumental activities of daily living (IADLs) that can be disrupted in Mild Cognitive Impairment (MCI) and are more significantly disrupted in Alzheimer's Disease (AD). In another study \cite{s6}, a predefined set of activities, including basic activities (such as walking or sitting) and IADLs (e.g., cooking, eating, or personal hygiene activities), reflecting money/self-management skills and travel/event memory abilities, were found to be most related to the sensor behavior data. Studies \cite{CV11} and \cite{cv12} used an automatic video monitoring system in a room to compare the motor abilities (e.g., walking speed) of a controlled healthy group and Alzheimer's disease patients with three designed tasks that consider different levels of autonomy (both walking and IADLs); statistically significant differences were observed. However, walking exercise in the room is limited to a short distance and time, which is less reliable \cite{CV11}. Another study \cite{cv10} monitored three instrumental activities, such as preparing the pillbox, preparing tea, and making/receiving a phone call, and the results showed that the system could distinguish between healthy and mild cognitive impairment patients based on task execution time. 

While designed tasks and environment modifications of usual behavior have their merits, the ideal strategy to capture functional decline accurately and reliably is to observe the daily behavior of individuals where they spend most of their time: at home \cite{s6}.

Most smart homes can monitor natural daily activities for the long term. A report \cite{report1} shows that functioning in five core Activities of Daily Living (ADLs) is typically used to describe the extent of chronic disability among the elderly. These core ADLs include (1) bathing; (2) dressing; (3) using the toilet; (4) transferring from bed to chair, and (5) feeding oneself. In the study \cite{s7}, the focus was on activities such as cooking, eating, sleeping, personal hygiene, taking medicine, working, leaving home, entering home, bathing, relaxing, bed-toilet transition, washing dishes, and other activities. In the work \cite{hard1}, activities monitored included bathroom use, sleep, eating, drinking, gait, falls, and others. However, these activities occur in different rooms, which increases the complexity of the monitoring system by requiring many sensors to be placed around. In the work \cite{s9}, five activities of daily living (ADLs) were used for monitoring, including relaxing activities, such as watching TV, reading, and napping, which typically take place in a single location other than the bedroom and are important for characterizing daily routines and assessing functional independence. A study \cite{cv2} shows that sedentary periods are common in older adults' daily lives and are related to their health status. This provides a good opportunity to monitor older adults' health conditions when they typically remain sitting at a fixed location at home with their upper-body visible and moving most of the time. These sitting and relaxing activities informed our decisions when choosing monitoring scenarios.

\textbf{Time Scale:}
Various time scales are employed for different monitoring purposes. In the study \cite{f2}, smart home sensor data are utilized, and a 1-week time window is selected to compare behaviors between windows, with the aim of detecting health events such as radiation treatment, insomnia, and falls. In the work \cite{hard1}, the sensors generate an event every 7 seconds when continuous motion is detected. Additionally, physiological parameters are calculated during sleep, time spent away from home, gait speed, stride length, and stride time on a 24-hour basis. In the study \cite{s6}, changing time-series statistics for each variable are computed using a sliding window of length 7 days. Each designed activity for participants in the study \cite{s5} takes an average of 4 minutes to complete, while the testing session for eight activities lasts approximately 1 hour. Romdhane et al. \cite{cv12} employ the same automatic video system to track healthy and Alzheimer's participants. Three tasks, considering different levels of autonomy, are designed, ranging from 10 minutes to 30 minutes. 

Only a few studies have investigated the appropriate time scale for monitoring performance. In one study, the researchers explored how the activity features extracted at different window sizes affected the performance of predicting standard clinical assessment scores \cite{s9}. They found that a window of 30 days was suitable for extracting features for this purpose. Another study used smartphone data and broke the continuous data into windows of certain durations, ranging from 1 to 16 seconds \cite{ttt1}. It was shown that shorter windows performed better in classifying activities. In a study by Johnson et al. \cite{s7}, different window sizes for the sensor events were considered, ranging from 10 to 150 events. However, the window was fixed based on the number of sensor events, not on actual time, which may not work well when the types of sensors or scenarios are changed. These studies focused on the time scale clue for detector performance, and significant variations among detectors, features, and window sizes have been observed for the best performance among different (or similar) conditions and across different subjects \cite{s9}. In our work, we explore various temporal windows for feature extraction, and we aggregate temporal windows to longer time spans to determine the optimal time scale for effectively classifying health states.

\section{Methods}
Our goal is to use camera data to automatically identify behavioral changes related to {the unobservable} in the health or weakness of an individual. 
To achieve this objective, we first capture data on daily behaviors of individuals using an RGB-D video camera (Section \ref{sec:dataC}). Next, we extract important behavioral features from this data (Section \ref{sec:featureE}). To infer the weakness states of an individual, we frame it as a classification problem (Section \ref{sec:healthC}). Following this, we employ multiple classifiers and information-based methods to rank the importance of the features and activities with respect to the weakness condition (Section \ref{sec:activityR}). Furthermore, we identify the optimal time scale for effectively observing behavioral changes by investigating various temporal windows (Section \ref{sec:timeS}). Finally, we model the dependencies among features using a Bayesian Network (Section \ref{sec:bayesM}). We build models for each individual and aggregate the results for a more comprehensive analysis of the impact of the weakness condition.

\subsection{Data Capture} \label{sec:dataC} 
Five healthy participants (age/gender/{handedness} 
of P1--P5: 35M / right, 35F / right, 25M / right, 60M / right, 25F / right) were monitored in designated room areas while engaged in five common daily activities. An RGB-D camera was positioned to observe the designated room areas, which were equipped with a chair, a desk, and other home items, as depicted in Fig. \ref{fig:scenario}. The five activities included: 1) reading a book, 2) using a personal computer (PC), 3) eating, 4) taking a nap, and 5) watching television (TV). Participants were predominantly seated while performing these activities, with their upper bodies visible most of the time. The camera facilitated the monitoring of both participants’ movements and their surroundings, extracting pertinent features. The camera was connected to a processor for real-time processing of captured visual data. 

Participants initiated monitoring by pressing start on the processor whenever they commenced a daily activity among the five listed activities. The camera then captured their behavior in real time throughout the monitoring period. Participants had the flexibility to cease an activity at any time, resulting in varying durations for each trial. The duration of activities ranged from minutes to hours based on individual preferences and circumstances. In order to label the activity being performed, the participants manually assigned an activity label to the monitoring period after each recording session, exclusively selected from the five available activities. Occasionally, two activities might coincide, such as eating while watching TV. In such cases, participants assigned only a single dominant activity label as preferred.

To simulate weakness, participants performed a workout, and their behavior was monitored for periods both before and after the workout to ascertain behavioral disparities. Since the fitness levels and exercise habits varied widely among participants, ranging from rarely exercising to more than two times regular gym training weekly, we asked participants to choose a workout based on their exercise level, whether weight training or cardio, to perform for up to 30 minutes until they subjectively felt tired. Similarly, depending on whether they had undergone a workout, participants assigned a health state label to each monitoring period, choosing from three options: ``normal'' (before the workout), ``same-day weakness'' (on the same day after the workout), and ``next-day weakness'' (on the next day after the workout) \footnote{As the volunteers knew whether or not they had exercised, there is probably some bias in the self-reported labels.}.
The monitoring process occurred in a naturalistic setting, devoid of scripting, interruptions, or intervention. 
To simulate a naturalistic setting rather than a designed task, participants were not bound by constraints; they had the liberty to skip activities or monitoring days from their daily routine. 
For instance, some participants never take a nap during the day, while others regularly exercise at night (i.e., no record of “same-day weakness”). So, we did not force them to carry out all activities, which is easily applicable to real-life monitoring and easier for participants to enter data. 
Consequently, the distribution of activity and health state labels is imbalanced, closely resembling natural daily routines. 
The distribution of these labels is depicted in Table \ref{tab:record}.

\textbf{Privacy}: To ensure personal privacy, only non-identifiable motion statistics and environmental contexts are stored as text logs. RGB frames and depth frames are processed in real-time and discarded immediately. The logs maintained the anonymity of the participants;  i.e., personally identifiable information, including names, faces, addresses, and other sensitive details, remained unrecorded.

This study received ethical approval from the School of Informatics Ethics Committee and individual agreements with the participants were made. 
Overall, a total of 260 activity monitoring records spanning 67 days were monitored for all the participants.

\begin{table}[]
\footnotesize
\caption{Statistics of the monitoring records from participants.
{Age, handedness, \& gender are reported in the main text.}
}
\begin{tabular}{cccc|cc|ccccc|ccccc}
\toprule
\multirow{2}{*}{ID} & \multirow{2}{*}{\begin{tabular}[c]{@{}c@{}}Monitor\\ Records\end{tabular}} & \multirow{2}{*}{\begin{tabular}[c]{@{}c@{}}Total\\ Hours\end{tabular}} & \multirow{2}{*}{\begin{tabular}[c]{@{}c@{}}Across\\ Days\end{tabular}} & \multicolumn{2}{c|}{Health (\%)} & \multicolumn{5}{c|}{Activity (\%)}    & \multicolumn{5}{c}{Mean Duration (minutes)}       \\
&      &    &       & normal       & weak  & 1.read  & 2.PC    & 3.eat   & 4.nap   & 5.watch & 1.read  & 2.PC    & 3.eat       & 4.nap  & 5.watch     \\
\midrule
P1    & 178     & 56.8   & 22    & 33.7       & 66.3  & 24.2  & 33.2  & 15.7  & 10.1  & 16.9  & 19.8   & 24.3   & 9.9   & 13.5   & 20      \\
P2  & 47  & 13.7   & 23   & 46.8       & 53.2      & 10.6  & 40.4  & -       & 12.8  & 36.2  & 13.8     & 20.7   & -   & 11   & 17.1    \\
P3  & 16  & 5.4  & 10   & 31   & 69  & 18.75 & 18.75 & 18.75 & 18.75 & 25    & 26     & 21.3    & 8.6   & 22.1   & 22.9     \\
P4  & 7  & 3    & 5    & 42.9       & 57.1      & 57.1  & -       & -       & -       & 42.9  & 23.3     & -     & -   & -    & 28.7    \\
P5  & 12  & 4.8  & 7    & 33.3       & 66.7      & 25    & 25    & 25    & -       & 25    & 29.8    & 22.8   & 7.4   & -    & 52.5    \\
\midrule
$\Sigma$  & 260  & 83.7  & 67    & 37.5       & 62.5      & 27.1    & 23.5    & 11.9    & 6.5       & 29.2    & 22.5    & 22.3   & 8.6   & 15.5    & 28.2    \\
\bottomrule
\end{tabular}
\label{tab:record}
\end{table}

\subsection{Feature Extraction} \label{sec:featureE}
Utilizing the rich visual information available, we extract 62 behavioral features, including aspects of human body movement and inactivity. Additionally, we examine four environmental contexts potentially related to an individual's behavior: objects, weather conditions, room lighting, and the time of day. 
For a comprehensive list of these features,  refer to Table \ref{tab:feautre}. 
To enhance the interpretability of the impact of health conditions, all features have semantic meaning. The objective is to identify effective descriptors for behavioral changes associated with health conditions, differentiating between normal and weakness states by discerning the significance of various features from the captured data. 

\textbf{Inactivity detection/ motion estimation:} The characteristics of movement and inactivity are crucial for health monitoring purposes, as they provide information on levels of physical activity, mobility, frailty, risk of falls, cognitive health, and more, as mentioned in Section \ref{sec:rw}. The inactivity detection algorithm involves several key steps. First, it employs nonparametric background modeling {in depth images}, then performs human region detection and nonhuman region suppression on the {depth foreground}, with pixel-level motion existence detection by RGB channels \cite{ourpaper1}. The period between two motion events (greater than or equal to 1 second) is detected as inactivity. This approach is sensitive enough to detect subtle movements such as finger movements and remains robust in various environmental lighting conditions, including low-light situations and the presence of television light. The method achieved a 0\% false positive rate with a ±3 frames temporal tolerance and a 3\% false negative rate for motion detection, as reported in \cite{ourpaper1}.

\begin{table}[h]
\footnotesize
\caption{List of extracted features from monitoring records.}
\begin{tabular}{lccc}
\toprule
    & Size   & \multicolumn{2}{l}{Description}    \\
    \midrule
Health (H)      & 1x2    & \multicolumn{2}{l}{normal, weakness}     \\
Activity (A)    & 1x5    & \multicolumn{2}{l}{read, PC, eat, nap, watch}  \\
1.   Lighting of room       & $R^1$     & \multicolumn{2}{l}{luma Y’601 of the captured scene}       \\
2. Time   of the day  & 1x3    & \multicolumn{2}{l}{6 am - 2 pm, 2 pm-10 pm, 10 pm-6 am}  \\
3.   Weather    & 1x2    & \multicolumn{2}{l}{suitable*   / not suitable (for outdoor activity)}  \\
4. Objects      & $R^{20}$    & \multicolumn{2}{l}{likelihood of presence of the 20 most frequent home objects}  \\
5. Duration of   activity   & $R^1$     & \multicolumn{2}{l}{seconds  (see main text for method)}  \\
6.   Ratio of inactivity    & $R^1$     & \multicolumn{2}{l}{\%}       \\
7. No.   of inactivity      & $R^1$     & \multicolumn{2}{l}{count of inactivity events (per min.) (see main text)}  \\
8. Duration of movement   & $R^1$     & \multicolumn{2}{l}{seconds  (see main text)}  \\
9.   Ratio of movement      & $R^1$     & \multicolumn{2}{l}{\%}       \\
10.   Movement pixel count  & $R^1$     & \multicolumn{2}{l}{$10^4$   pixels/s (see main text)}       \\
12.   Density of movement   & $R^1$     & \multicolumn{2}{l}{\% (movement pixel count / human pixel count)}      \\
14.   Scale/Spread of movement    & $R^1$     & \multicolumn{2}{l}{\%   (movement pixels bbox size/ human bbox size)}  \\
16.   Mean speed of movement      & $R^1$     & \multicolumn{2}{l}{pixel/s}  \\
19.   Distance of movement  & $R^1$     & \multicolumn{2}{l}{pixel/5min.}    \\
20-23.   Mean speed quartiles     & $R^1$     & \multicolumn{2}{l}{Q1 - Q4}  \\
28. STD   of speed    & $R^1$     & \multicolumn{2}{l}{pixel/s}  \\
    \midrule
\begin{tabular}[c]{@{}l@{}}Fine-grained   features ($R^1$)\\  (See footnotes   for explanation.)\end{tabular} & \begin{tabular}[c]{@{}l@{}}11. 10   in MPO*\\   13. 12   in MPO\\   15. 14 in   MPO\\  17. 16   in MPO\\    18. 28   in MPO\\   24 –   27. 20 – 23 in MPO\end{tabular} & \begin{tabular}[c]{@{}l@{}}44 – 46. 12 in TRL*\\   35 –   37. 12 in TRL-MPO\\   41 –   43. 14 in TRL\\ 32 –   34. 14 in TRL-MPO\\   38 –   40. 16 in TRL\\  29 –   31. 16 in TRL-MPO\end{tabular} & \begin{tabular}[c]{@{}l@{}}47 –   56. inactivity duration distribution (\%)\\ Ranges   (seconds): {[}1,2), {[}2,5), {[}5,10), {[}10,30), \\  {[}30,60), {[}60,), {[}2,), {[}5,), {[}10,),   {[}30,)\\    \\  57 –   66. movement duration distribution (\%)\\  Ranges   (seconds): {[}1,2), {[}2,5), {[}5,10), {[}10,30), \\  {[}30,60), {[}60,), {[}2,), {[}5,), {[}10,),   {[}30,)\end{tabular} \\
\bottomrule
\end{tabular}
\label{tab:feautre}
\\
\footnotesize MPO* (Movement Period Only) refers to statistics computed only from the period during which  movement is considered, excluding data from inactive periods. 
For instance, the mean speed during MPO is typically higher compared to the mean speed over the entire video duration because there are instances when the person is not moving. TRL* (Top, Right, and Left of body regions) involves dividing the detected human region (bounding box) into 2x2 sub-boxes. The Top body region contains sub-boxes 1 and 2, the Right body region contains sub-boxes 1 and 3, and the Left body region contains sub-boxes 2 and 4. Suitable* weather conditions are selected from a list of 27 weather types. The "suitable" conditions include Clear, Fair, Cloudy, and Overcast, while the other 23 weather types are considered "not suitable" for outdoor activity, including Fog, Rain, Sleet, Snowfall, Storm, Hail, and so on. (See full list at \cite{weaapi}).
\end{table}

Once the moving pixels are extracted, a dense 2D optical flow estimator \cite{opticalflow} is applied to them.
This optical flow analysis helps describe the intensity of the body movement, in terms of 2D velocity. This is advantageous for capturing small movements in the upper body, such as those involving the head and hands, which can be significant in daily upper-body activities. 
Then, various statistics are derived from the data to provide valuable features. These include metrics like inactivity count, inactivity duration, movement distance, standard deviation, movement distribution, and interquartile range (IQR). 

The combination of inactivity detection with optical flow is particularly beneficial because it can effectively represent small movements at the pixel-level, and simultaneously ensures the accuracy of pixels with true motion, ignoring the statistical noise from the dense optical flow estimation. This makes it suitable for activities involving numerous fine motor movements. Additionally, one advantage of employing a fixed camera for monitoring a stationary area is it ensures a relatively consistent distance between the camera and the person being monitored. This consistency makes 2D optical flow magnitude a reliable estimator of motion intensity. 
This contrasts with the potential inaccuracies associated with estimating 3D joint coordinates from noisy depth measurements, as well as the challenges of detecting key joints in occlusion situations, for subtle movements, and in low-lighting environments. 
{We evaluated the accuracy of 2D human body movement speed estimation using several datasets. An optical flow-based method combined with true motion detection achieved better results than pose estimation methods or using optical flow alone, with a Root Mean Square Error (RMSE) of 0–2.4 pixels per frame for 2D human body movement speed estimation}~\cite{opph}.

\textbf{Fine-grained descriptors:} 
Fine-grained descriptors allow additional insights into the spatial density and scale of movement, enhancing our ability to characterize the range and intensity of body movements. The density of movement is 
the proportion of movement pixels to the total number of pixels within the body region. Conversely, the scale of movement is the ratio between the size of the movement bounding box and the size of the human bounding box. Furthermore, we extract features from different body regions, including the Top, Right, and Left sides (TRL) of the detected human body. 
This approach enables us to focus on parts of the body separately. 
Additionally, we employ a fine-grained temporal descriptor that emphasizes statistics derived specifically from the periods of movement only (MPO), rather than considering the entire period. 
Further, the inactivity and motion durations are separated into multiple intervals to compute their detailed distributions. 
See Table \ref{tab:feautre} for details of the fine-grained descriptors. 
These fine-grained descriptors help to develop deeper insights into the nature of the movements. 
For instance, 
it helps 
where a single hand movement dominates the activity or where head movement takes precedence. It can also highlight significant temporal pauses within an activity.

\textbf{Environmental context:} The environmental context is also an essential aspect of the monitoring system. We employ a pre-trained YOLOv5 model \cite{yolov5} capable of detecting common objects, including humans. The model exhibits a mean Average Precision (mAP) of 56.8\% on the COCO val2017 dataset \cite{coco}. Real-time local weather information is accessed through an online Weather API \cite{weaapi}. The room illumination is calculated from the image by referencing the Rec. 601 luma standard \cite{601}. Time of the day is divided into three 8-hour segments, enabling us to capture temporal variations in daily routines. These environmental context features enrich our understanding of the monitored person.

\begin{figure}[h]
    \centering
    \footnotesize
    \stackunder[5pt]{\includegraphics[height=0.234\linewidth]{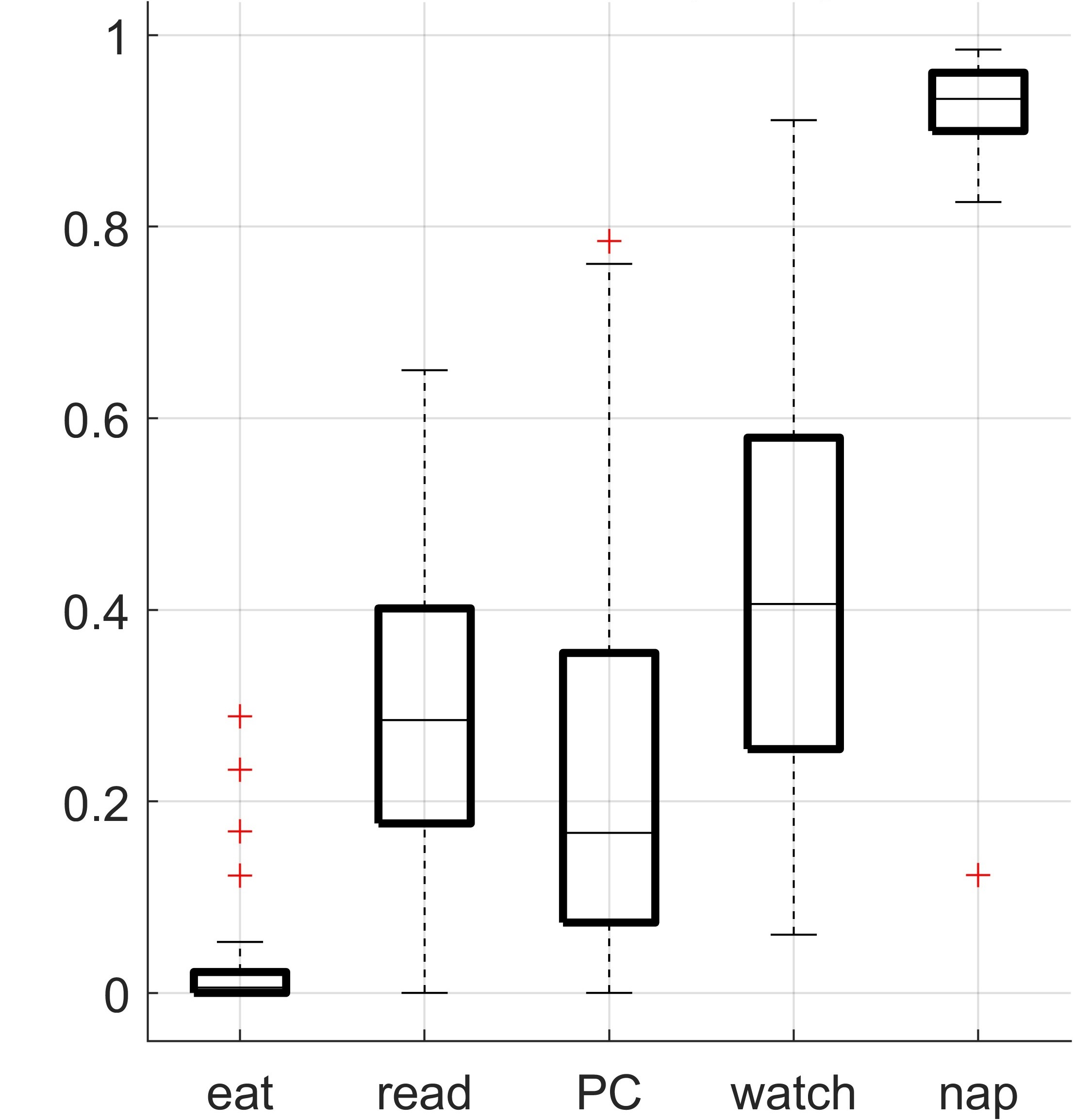}}{Inactivity ratio (Act.)} \ \ 
    \stackunder[5pt]{\includegraphics[height=0.234\linewidth]{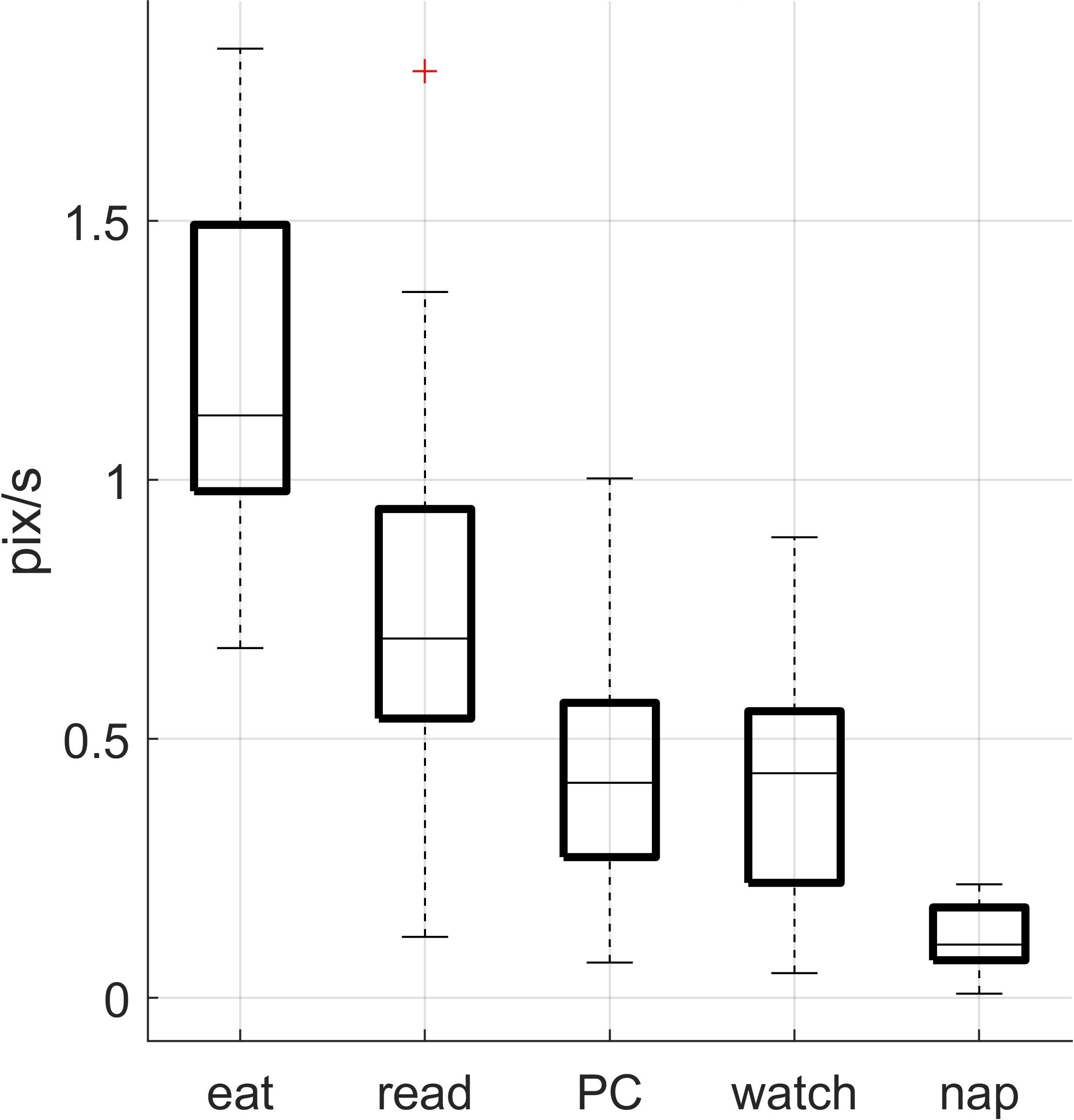}}{Left body speed (Act.)} \ \ 
    \stackunder[5pt]{\includegraphics[height=0.234\linewidth]{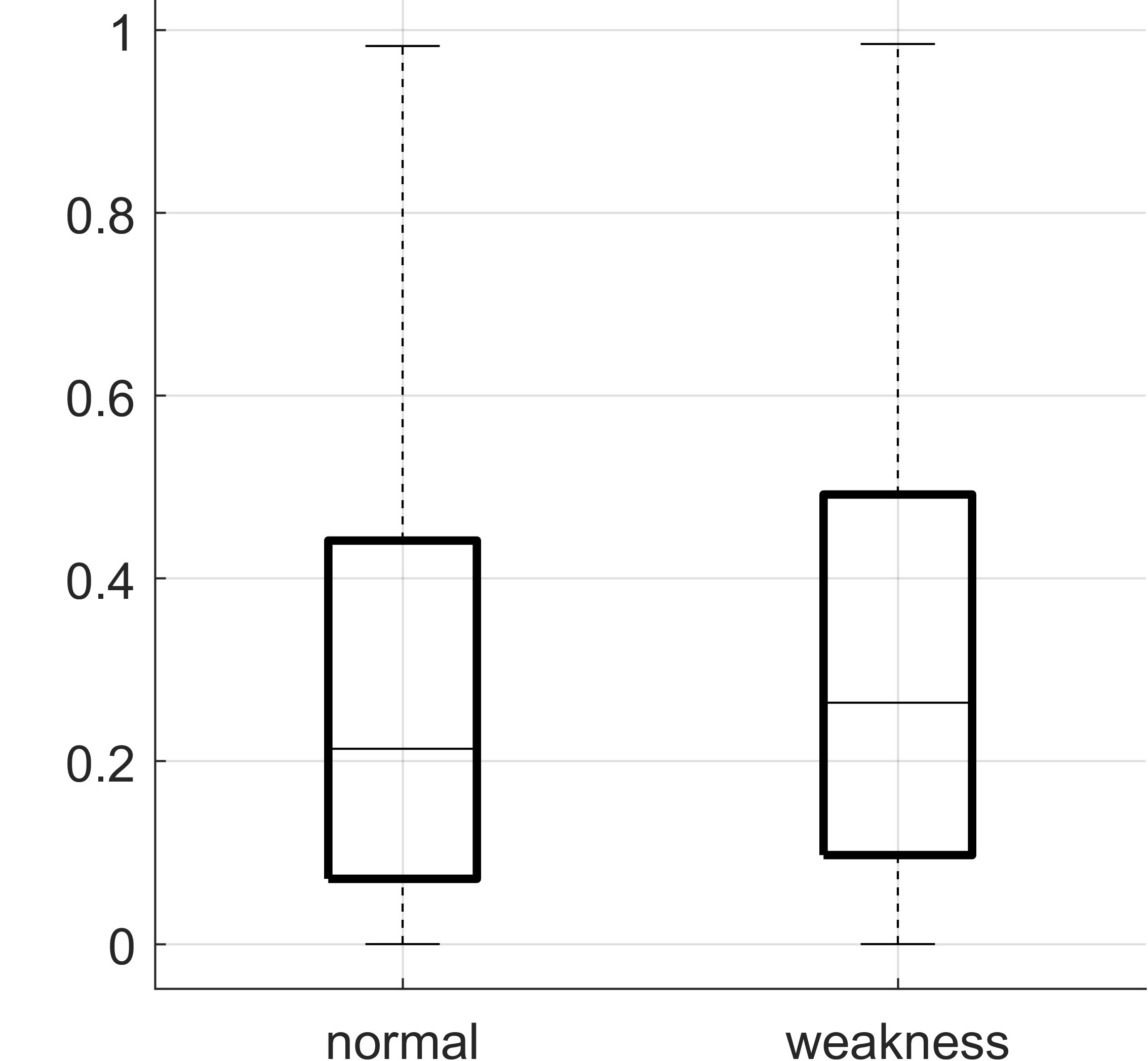}}{Inactivity ratio (Hea.)} \ \  
    \stackunder[5pt]{\includegraphics[height=0.234\linewidth]{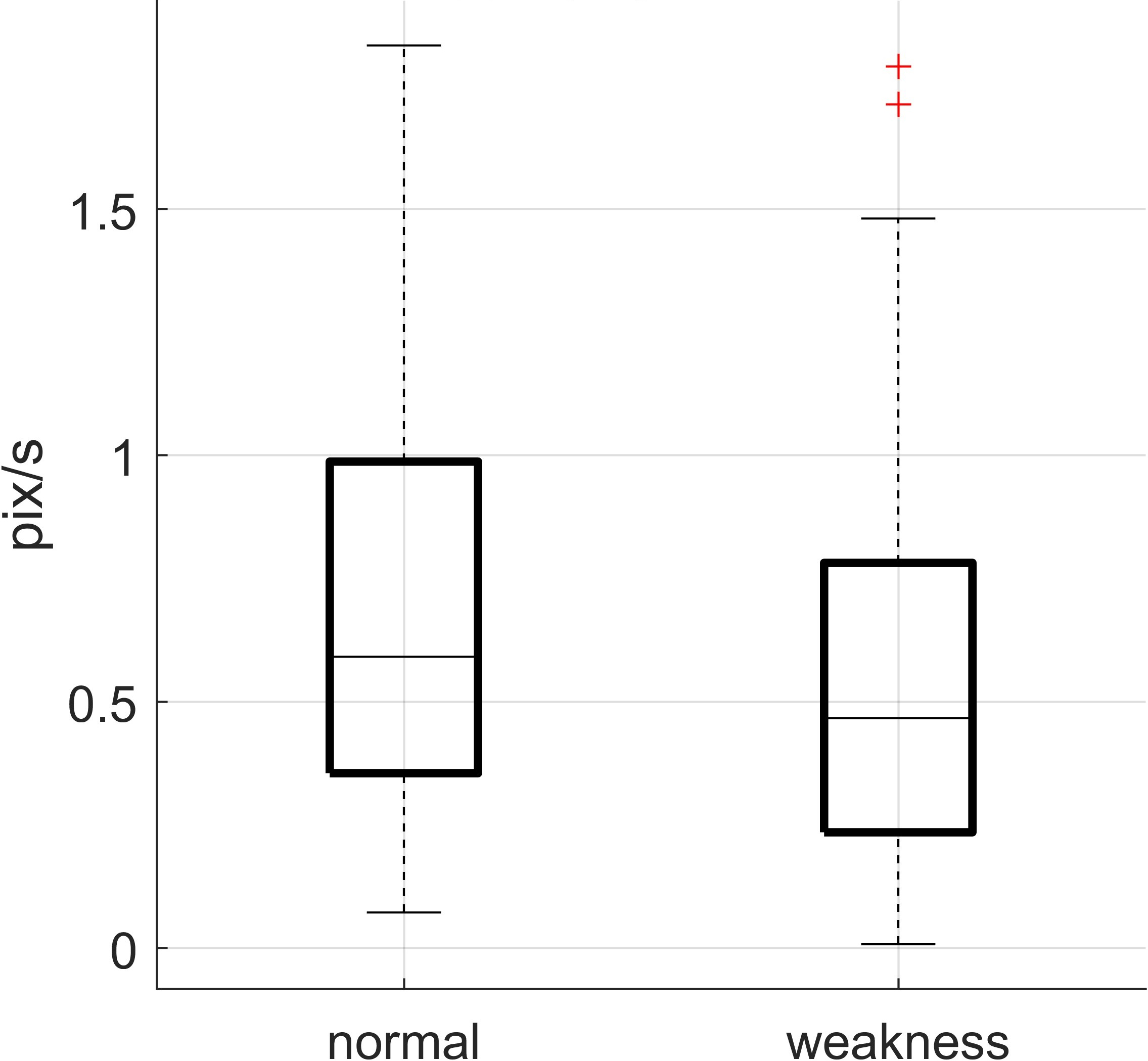}}{Left body speed (Hea.)} 
    \caption{Example of two features that differ among activities and between health states for a participant in all activities.
    {The left two plots show how the feature values vary according to the activity (Act.). The right two plots show how the feature values vary (slightly) between the Normal and Weakness health states (Hea.).}
    }
    \label{fig:example}
\end{figure}

\subsection{Health State Classification} \label{sec:healthC}
Our initial inquiry revolves around the ability to distinguish patterns associated with weakness from typical behaviors using the measured features. 
In this context, this is a classification problem, with the goal of discerning between states of normal health and instances of weakness. Given the substantial variability in the duration of post-workout weakness, which can manifest and dissipate within 72 
hours\footnote{The duration of post-workout weakness can depend on various factors, including an individual's fitness level, the intensity of the workout, the type of exercise performed, and overall health. Generally, post-workout weakness may be experienced immediately after an intense exercise session and can persist for several hours to a couple of days \cite{DOMS}.}, here we have merged the categories of ``same-day weakness'' and ``next-day weakness'' into a unified ``weakness'' label, which is distinct from the label signifying ``normal'' health status (i.e., pre-workout).

For each monitoring record, we denote the health states as $\mathcal{H} = \{H_1, H_2\}$, the activities as $\mathcal{A} =\{A_1, A_2, ..., A_5\}$, and the extracted features {(i.e., behavioral and environmental)} as $\mathcal{F} = \{F_1, F_2, ..., F_{66}\}$. {Then the record data can be denoted as $\mathcal{D} =\{(H,A,F) \mid H \in \mathcal{H}, A \in \mathcal{A}, F \in \mathcal{F}\}$}. 
The goal is to identify the combination of these elements that maximizes the performance of the classifiers for health state classification. 
The optimal set of features and activities can be succinctly expressed as:


\begin{equation} \label{eq:select}
(S_A^*,S_F^*) = argmax_{d \in \mathcal{D}{(S_A^*,S_F^*)}} \text{HSA}(H_d \mid A_d, F_d) 
\end{equation}
\ \\

\noindent
where {HSA() estimates the health state classification accuracy,} $S_A^*$ is a subset of $\mathcal{A}$,  $S_F^*$ is a subset of $\mathcal{F}$, {and $\mathcal{D}{(S_A^*,S_F^*)}$ is a subset of $\mathcal{D}$. We require the selected data size, $n = |\mathcal{D}{(S_A^*,S_F^*)}|$ to satisfy $n > 0.4 n_{\mathcal{D}}$, where $n_{\mathcal{D}}  = |\mathcal{D}|$. 
This ensures that the selected samples are adequately representative despite the large imbalance in activity types within the monitoring data.}

To address this classification task, we examine all possible combinations of activities performed by the participants. 
This amounts to a total of $2^m$ combinations, where $m$ is the number of activity types monitored for a participant. 
{Although it would be ideal to find a single, generalized set of activities and features that allow reliable health state classification for all current participants (and future users as well), limitations in the current data require an alternative approach. We instead investigate the possibility of estimating individual health states through personalized sets of features and activities. This also makes practical sense because not all people would do the same activities, and not all people would respond in the same way to a given activity. Healthcare applications tailored to each individual are a growing trend \cite{kosorok2019precision}.}
For each combination of activities, we gradually introduce features for training and testing the classifiers, following a forward selection approach. The results of classifying normal and weakness states with three different classifiers are presented in Fig. \ref{fig:acc}.

\begin{table}[h]
\footnotesize
\caption{Feature ranking and Activity ranking methods.
{The top block lists classifiers used for feature ranking evaluations, the middle block lists information-based feature ranking methods, and the bottom block gives the rank aggregation methods used.}
}
\begin{tabularx}{\textwidth}{l|X|X}
\toprule
Method  & Description   & Score Calculator       \\
\midrule
SVM \cite{svm}  & Support Vector Machine: find the optimal decision boundary that maximizes the margin between the classes.   &  \multirow{5}{*}{$s_i = \text{accuracy of classifying health states } H \text{ by feature } F_i$}  \\
BN \cite{friedman1997bayesian} & Bayesian Network: model the probabilistic dependencies between the features and the target class.     &      \\
RF \cite{breiman2001random} & Random Forests: build an ensemble of decision trees to improve generalization and reduce overfitting.   &      \\
\midrule
FDR \cite{fisher1936use}  & Fisher Discriminate Ratio: emphasize the discriminative power of each feature in distinguishing between the classes.  &  $s_i = {(\mu^{(H_1)}(F_i) - \mu^{(H_2)}(F_i))^2}\slash{(\sigma^{(H_1)}(F_i) + \sigma^{(H_2)}(F_i))}$ \\
MI \cite{cover1999elements} & Mutual Information: measure the amount of information that a feature provides for class prediction.     & $s_i =  P(F_i, H_1) \log  \frac{P(F_i, H_1)}{P(F_i) P(H_1)}  + P(F_i, H_2) \log  \frac{P(F_i, H_2)}{P(F_i) P(H_2)} $ \\ 
CFS \cite{hall1999correlation} & Correlation-based Feature Selection: consider the relevance and redundancy of features in relation to the target class and each other.  & 
{$s_i =   {\text{abs}(\text{corr}}(F_i, H)  \slash  \text{mean}(\text{corr} (F_{i}, F_{j})))$}
\\
\midrule
BC \cite{emerson2013original} & Borda Count: aggregates the scores by assigning points to each feature based on its rank from each method and summing up the points to obtain the final ranking. & $R_{agg} = \sum_{k=1}^K (n - r_k + 1)$, where $n$ is the number of features and $r_k$ is the rank of the feature according to method $k$.  \\
NWA \cite{cochran1977sampling} & Normalization and Weighted Average: the scores from each method are normalized to a common scale and then combined using weighted averaging.    &  $R_{agg} = \sum_{k=1}^K w_k \cdot \text{normalize}(s_k)\slash{\sum_{k=1}^K w_k}$,   where $w_k$ is the weight assigned to method k and normalize($s_k$) is the normalized score of method $k$.  \\
Cb \cite{woehr2015justifying} & Consensus-based: find a consensus among the rankings from different methods.  &  $R_{agg} = \text{mode}(r_1^{1:m}, r_2^{1:m}, ..., r_k^{1:m})$,   where $mode()$ returns the most frequent features that appear among the top rankings (with top-$m$ features for activity ranking). \\
\bottomrule 
\end{tabularx}
\label{tab:aggmeth}
\end{table}

\subsection{Feature and Activity Ranking} \label{sec:activityR}
After extracting features from participants' daily common activities considering both their behavioral and environmental context, our second objective is to identify the most relevant features and activities that are effective indicators of the health states. To accomplish this, we rank activities and features based on their capacity to distinguish weakness from normal states. To ensure the stability and consistency of these rankings, multiple methods are typically employed \cite{rankmeth}.

In this context, we first explore the relationship between the i-th feature, denoted as $F_i \in \mathcal{F}$, and the health state variable $H$. To rank these features, we utilize a ranking method referred to as $M$, which provides a score $s_i$ defined as:
\begin{equation}
     s_i = M(F_i, H).
\end{equation}
  
The scores can be obtained using various approaches. Once we have calculated the scores for all features, we sort them to derive the rank of all features: 
\begin{equation}
\mathcal{R} = \mathrm{sort}(\mathcal{S}), \quad \text{where } \mathcal{S} = \{s_1, s_2, \ldots, s_{|\mathcal{F}|}\}. 
\end{equation}

However, different ranking methods often take into account specific properties of the features, resulting in inconsistent results. Here, we utilize a diverse set of techniques for feature ranking, including three classifiers, Bayesian Network (BN), Random Forest (RF), and Support Vector Machine (SVM), and three information-based methods, Fisher Discriminate Ratio (FDR), Mutual Information (MI), and Correlation-based Feature Selection (CFS). Each of these basic ranking methods considers distinct aspects of the features, as outlined in Table \ref{tab:aggmeth}.

Considering scores derived from the $k$th basic method as $\mathcal{S}_k$ and the rank derived from this method as $\mathcal{R}_k$, we use aggregation techniques to integrate the ranking results from all the above basic methods. The three aggregate approaches include Borda Count (BC), Normalization and Weighted Average (NWA), and Consensus-based (Cb). Each of these aggregation approaches combines the score derived from the individual methods. 

A final score $S_\mathrm{agg}$ that represents the overall importance of the features for the health states is derived as an unweighted sum of normalized scores:
\begin{equation}         \label{eq:aggscore}   
S_\mathrm{agg} = \mathrm{normalize}(\mathrm{BC} (\mathcal{R}_1, \mathcal{R}_2, \ldots, \mathcal{R}_k)) +  \mathrm{normalize}(\mathrm{NWA} (\mathcal{S}_1, \mathcal{S}_2, \ldots, \mathcal{S}_k)) + \mathrm{normalize}(\mathrm{Cb} (\mathcal{R}_1, \mathcal{R}_2, \ldots, \mathcal{R}_k)). 
\end{equation}

The same pipeline is employed for activity ranking. The importance scores for each activity are determined by examining the top features within each activity that demonstrate the ability to distinguish between the health states. By leveraging a diverse range of techniques, we aim to provide a comprehensive and accurate ranking of the most significant factors contributing to the identification of weakness health states. The top-ranked activities and features for each participant are shown in Fig. \ref{fig:aheach} and Fig. \ref{fig:fheach} respectively.
{
The figures show that detection of weakness in every person is possible, but is best characterized by a different set of features and activities for each participant.}

\subsection{Optimal Time Scale} \label{sec:timeS}
Normal activities vary in duration, ranging from minutes to hours. Therefore, it is crucial to determine the optimal time scale for effectively monitoring statistics related to health states during these activities. By systematically analyzing different time windows,  the time scales that yield the most informative and discriminative features can be estimated, thereby enhancing the precision of the health state classification and behavioral change detection.

Different temporal windows for feature extraction, and different time spans for aggregating the results from temporal windows, are considered with durations ranging from  
\{ 30s, 60s, 120s, 300s, 600s, 1200s \}.
Short windows (i.e., < 30s)  increase the calculation burden and make it hard to capture longer-term characteristics, while too long windows (> 20 minutes) may mix different activities together.
For the longer time scales, infrequent activities may have too few samples to train the classifiers effectively. 

Subsequently, we aggregate the results obtained from temporal windows to longer time spans, such as 8-hours (refer to the "time-of-day" feature in Table \ref{tab:feautre}) and a full-day, for a more robust result, assuming that the health state stays consistent during the time span. 
The effectiveness of temporal windows for classifying health states is illustrated in Fig. \ref{fig:timescale}.

   \subsection{Modeling with Bayesian Network} \label{sec:bayesM}
A Bayesian Network is used to model the relationship between health states, environmental factors, behaviors, and activity types. 
This network captures the probabilistic dependencies among these variables, enabling inferences and predictions about the most likely health states based on observations. 
Furthermore, the learned model facilitates causal inference, allowing identification of the health-related features that are most likely to influence health outcomes.

\begin{figure}[h]
    \centering
    \includegraphics[width=0.39\linewidth]{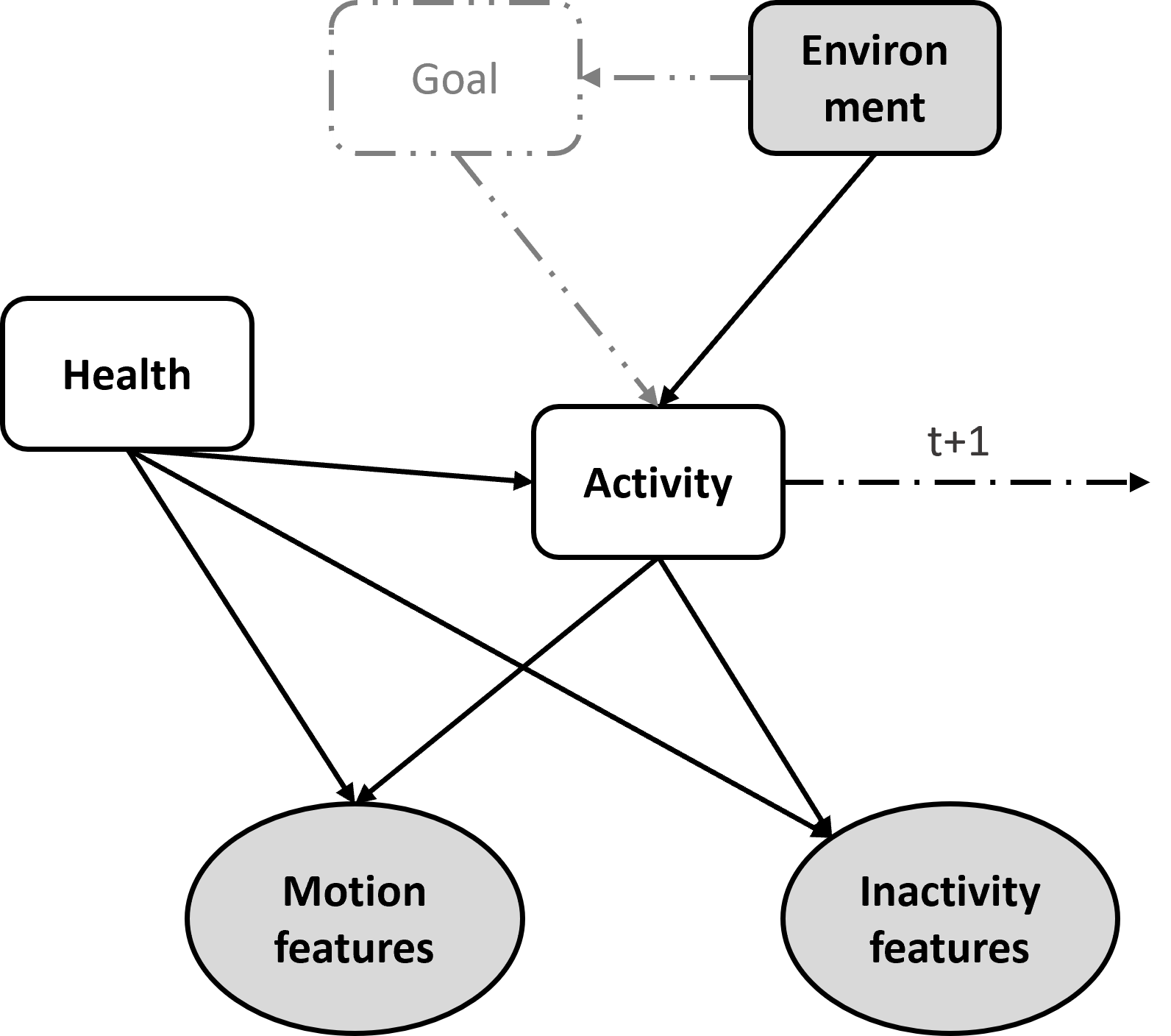}
    \caption{Structure of the Bayesian Network. Continuous variables are circles and discrete variables are rectangles. Shaded nodes represent observable features. The unobservable `Goal' node is omitted. The network can be extended to a dynamic model by adding temporal links. }
    \label{fig:BN}
\end{figure}

The structure of the Bayesian Network is depicted in Fig. \ref{fig:BN}. 
This model incorporates health states and behavioral features into the classic Goal-Environment-Activity model. The dependencies of the observed variables are as follows: 

"Activity" is influenced by both the individual's "health state" and the "environment". For example, when an individual feels weak, they may tend to choose less intense activity types. 
Additionally, different environmental factors, such as the presence of specific objects or the time of day, can also impact on the selection of activities. 
For instance, the presence of specific objects may be a good clue to indicate an individual is engaged in eating or working on the PC. 
Certain activities may be preferred at specific times of the day. Weather conditions, such as good weather, may make individuals more likely to engage in outdoor activities, while they may opt for indoor activities during rainy or cold weather. 

The individual's "behavior" is dependent on both the "activity type" and their "health state". 
For example, when feeling weak, the individual's movements may be slower, and they may take more rest time. On the other hand, when performing activities such as napping or working on a PC, the frequency and pattern of their movements may differ significantly. Fig. \ref{fig:example} shows an example of how behavioral features differ among different activities.

The "goal" (i.e., user intention) is considered unobservable and is omitted from the network. While it may play a role in influencing activity choices and behavior, it is not directly observed or measured in this context. 
The "health state" and the "environment" are considered independent of each other in the model. This means that changes in the environment do not directly influence the individual's health state, and vice versa. 

Based on the dependencies described above, the Bayesian Network equation for the health state ($H$), environment ($E$), activity type ($A$), and features ($F$) can be written as follows: 
\begin{equation}
    P(H, E, A, F) = P(H) \cdot P(E) \cdot P(A \mid H, E) \cdot P(F \mid A, H),
\end{equation}    
where $P(H)$ represents the probability distribution of the health state, $P(E)$ represents the probability distribution of the environment, $P(A \mid H, E)$ represents the conditional probability of the activity type given the health state and the environment. $P(F \mid A, H)$ represents the conditional probability of the features given the activity type and the health state, as:
\begin{equation}
 P(F \mid A,H) = P(mf \mid A,H) \cdot P(if \mid A,H),
 \end{equation}  
where $mf$ and $if$ are motion features and inactivity features, respectively. 

To infer the health state (H) given the observed variables, we can use Bayes' theorem. Bayes' theorem allows us to update our beliefs about the health state based on the observed evidence. The equation for performing inference in this case is as follows:
\begin{equation}
P(H \mid A, E, F) = {P(H, E, A, F)}/{P(A, E, F)} 
= {P(H)\cdot P(E) \cdot P(A \mid H, E) \cdot P(F \mid A, H)}/{P(A, E, F)},
\end{equation}  
where $P(H \mid A, E, F)$ represents the posterior probability of the health state given the observed activity type, environment, and behavioral features, and $P(A, E, F)$ is the evidence, which is calculated as the sum of the joint probabilities of all possible health states as $P(A, E, F) = \sum_{H} P(H) \cdot P(E) \cdot P(A \mid H, E) \cdot P(F \mid A, H)$.

If the activity type is unknown on some occasions it also needs to be inferred:
\begin{equation}
P(A, H \mid E, F) =  {P(H) \cdot P(E) \cdot P(A \mid H, E) \cdot P(F \mid A, H)} / {P(E, F)},
\end{equation} 
where $ P(E, F) = \sum_{A,H} P(H) \cdot P(E) \cdot P(A \mid H, E) \cdot P(F \mid A, H) $.

\section{Results}


\subsection{Experimental Details} \label{sec:im}

\textbf{Health state classification }
For health state classification, the training data is drawn from the monitoring records of each participant (details are shown in Table \ref{tab:record}). 
To achieve optimal performance, activities, and features are iteratively added to the classifiers using a forward selection approach. 
At each stage of the iterative process, {all classifiers were trained and tested using 5-fold cross-validation with shuffled samples, where 4 folds were used for training and 1 for validation.}
For the Bayesian network, the structure is predefined as shown in Fig. \ref{fig:BN}. 
The size of the nodes is based on the sizes of the selected features, as shown in Table \ref{tab:feautre}, and the initial probabilities of all nodes are randomly set. 
The parameters of each adjustable node are then set to their ML/MAP values using batch EM \cite{bayes}. For the SVM, a two-class SVM is used to classify normal and weakness health states. For the RF, a parameter of 100 trees is chosen.
For the CNN-GRU, one convolution layer (kernel size 3) and one GRU layer (hidden units 128) are used. The initial learning rate is set to 0.001, and the model is trained for 100 epochs. For the LSTM, one layer with 120 hidden units is used. The hyperparameters are the same as those used for the CNN-GRU. The models with the best validation loss are selected.

Due to the inherent imbalance in activity classes, the training process prioritized F1-macro as the primary evaluation metric, aiming for robust performance across all activity classes.  In each training loop, the feature or activity with the highest F1-macro score was added to the classifiers. This process continued until a predetermined number of features (30 in this case) were included and all activity combinations were represented. Finally, the best performing model was selected, {typically after less than 19 features were added.}
To minimize results bias caused by a lack of data from infrequently performed activities, we only selected combinations of activities 
that cover over 40\% of the training data (Equation \ref{eq:select}) for optimal classification performance. 
This approach ensures that the classification model is trained on a representative sample of activities, preventing it from being overly influenced by data from a single activity (top-1).

\textbf{Optimal time scale }
To determine the optimal timescale, each monitoring record is segmented into temporal windows of fixed length (ranging from 30 seconds to 1200 seconds), and the features within each temporal window are utilized for classification. 
The classification labels obtained from each window are aggregated to a record-level label using majority voting. 
The record-level labels are further aggregated to 8-hour and daily levels based on their timestamps, also using majority voting.

\begin{figure}
    \centering
    \includegraphics[width=0.8\linewidth]{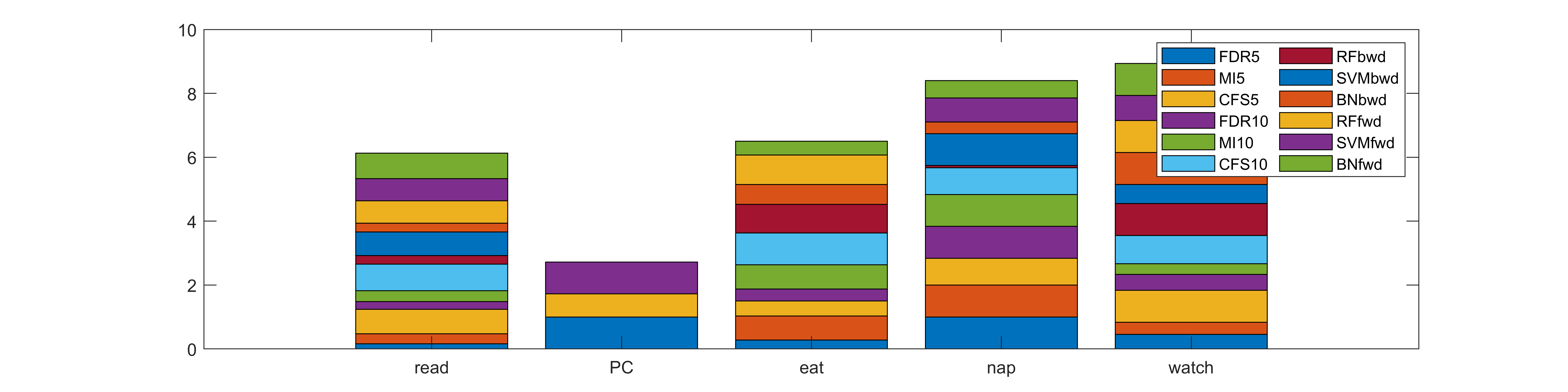}
    \caption{Example of activity ranking results for five activities (horizontal axis) of a participant (P1) using the basic methods (before aggregation). 
    {The different colored rectangles refer to the 12 different metrics for ranking.}
    The top 5 and 10 features were used for information-based methods (FDR, MI, CFS), while forward and backward selection methods were applied for classifiers (SVM, BN, RF). (See Table \ref{tab:aggmeth} for details.) The height of each rectangle is the normalized score (0-1) of that metric for that activity. The vertical axis represents cumulative scores from all metrics for each activity type. Variations in scores across metrics highlight the need for aggregation to achieve a more reliable ranking. {Activities with consistently high scores across metrics likely serve as better indicators of the health states.}
    }
    \label{fig:actrank}
\end{figure}

\textbf{Feature and activity ranking }
For ranking features and activities, the same set of basic classifiers (BN, SVM, RF) are utilized. 
Both forward and backward selection are applied to each activity to assess the performance of the classifiers. 
For each information-based ranking method (FDR, MI, CFS), the top 5 and 10 features are selected for consideration. 
The ranking scores generated by different basic methods are illustrated in Fig. \ref{fig:actrank}. Different basic ranking methods yield slightly different scores, indicating that a single method lacks robustness for activity ranking due to inconsistencies in the results from multiple methods. Consequently, aggregating different methods can lead to a more reliable ranking.

Subsequently, aggregation methods (BC, NWA, Cb) are utilized to derive the aggregated scores from the aforementioned six methods. Then, all scores from different aggregation methods are normalized to a range between zero and one and then summed to obtain the final score (Equation \ref{eq:aggscore}).

\subsection{Health State Classification Results} \label{sec:class}
\begin{figure}[h]
    \centering
    \includegraphics[width=0.5\linewidth]{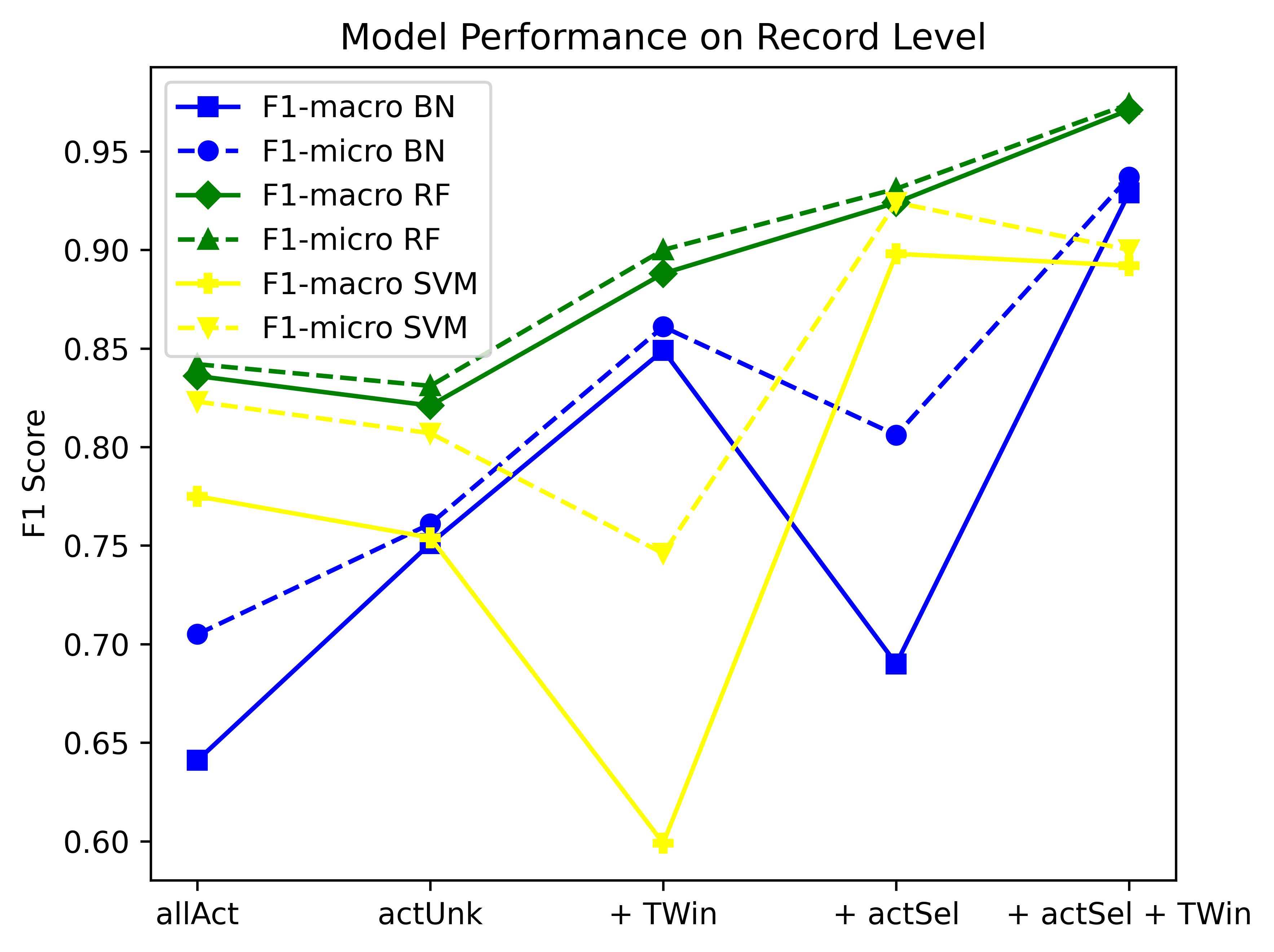}
    \caption{F1 score for inferring health states on all monitoring records with three classifiers (Bayesian Network (BN), Random Forest (RF), and Supportive Vector Machine (SVM)). allAct: classify with all activities; actUnk: classify with all activities with activity labels unknown; + TWin: classify with all activities with optimal temporal windows for feature extraction; + actSel: classify with optimally selected combination of activities (>40\% data coverage, details are given in Section \ref{sec:im} Implementation Details.); + actSel + TWin: classify with optimally selected combination of activities and optimal temporal windows for feature extraction.}
    \label{fig:acc}
\end{figure}

Fig. \ref{fig:acc} presents the results for classifying normal and weakness health states at the record-level, with each sample representing a complete monitoring record corresponding to an entire activity duration. When training our models with all activities at the record-level, we achieved average accuracy (F1-micro) of 0.71, 0.84, and 0.82 for BN, RF, and SVM, respectively, distinguishing between normal and weakness among all participants. By exploring different temporal windows for feature extraction, we observed an improvement in the average accuracy of the three classifiers to 0.84 when considering all activities across all participants. Upon selecting representative activities by assessing all possible activity combinations for each participant, we identified certain combinations of activities\footnote{Selected combinations of activities must have over 40\% data coverage. Details are given in Section \ref{sec:im} Implementation Details.} that yielded an average of 0.89 for the three classifiers among all participants. Furthermore, by carefully choosing both the optimal activity combinations and timescales for feature extraction, we achieved an enhanced accuracy of 0.94 for the three classifiers. 

\subsection{Optimal Time Scale Results\label{sec:timescale}}
\begin{figure}[h]
    \centering
    \footnotesize
    \stackunder[5pt]{\includegraphics[height=0.22\linewidth]{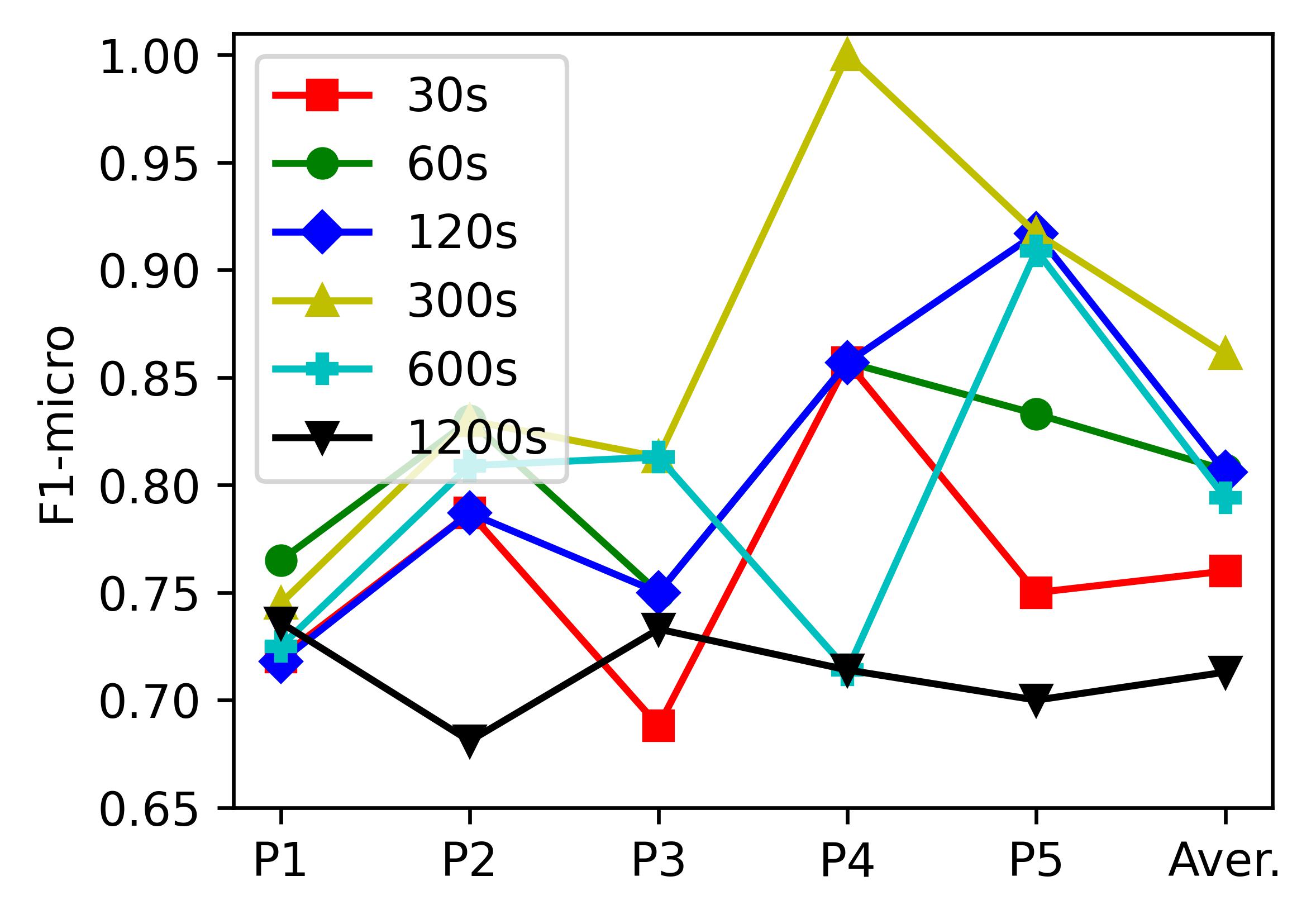}}{Record-level}
    \stackunder[5pt]{\includegraphics[height=0.22\linewidth]{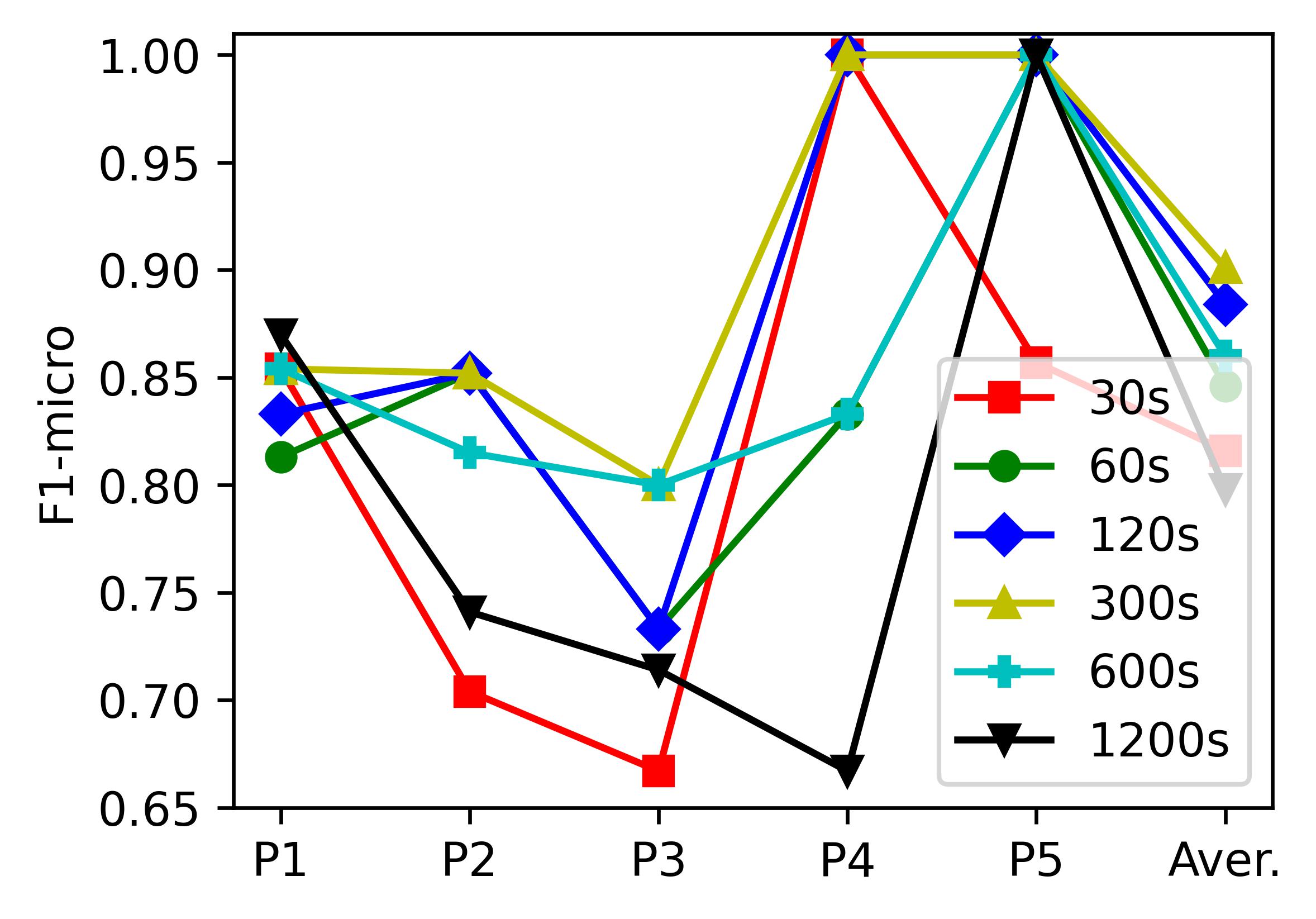}}{8-hour level}
    \stackunder[5pt]{\includegraphics[height=0.22\linewidth]{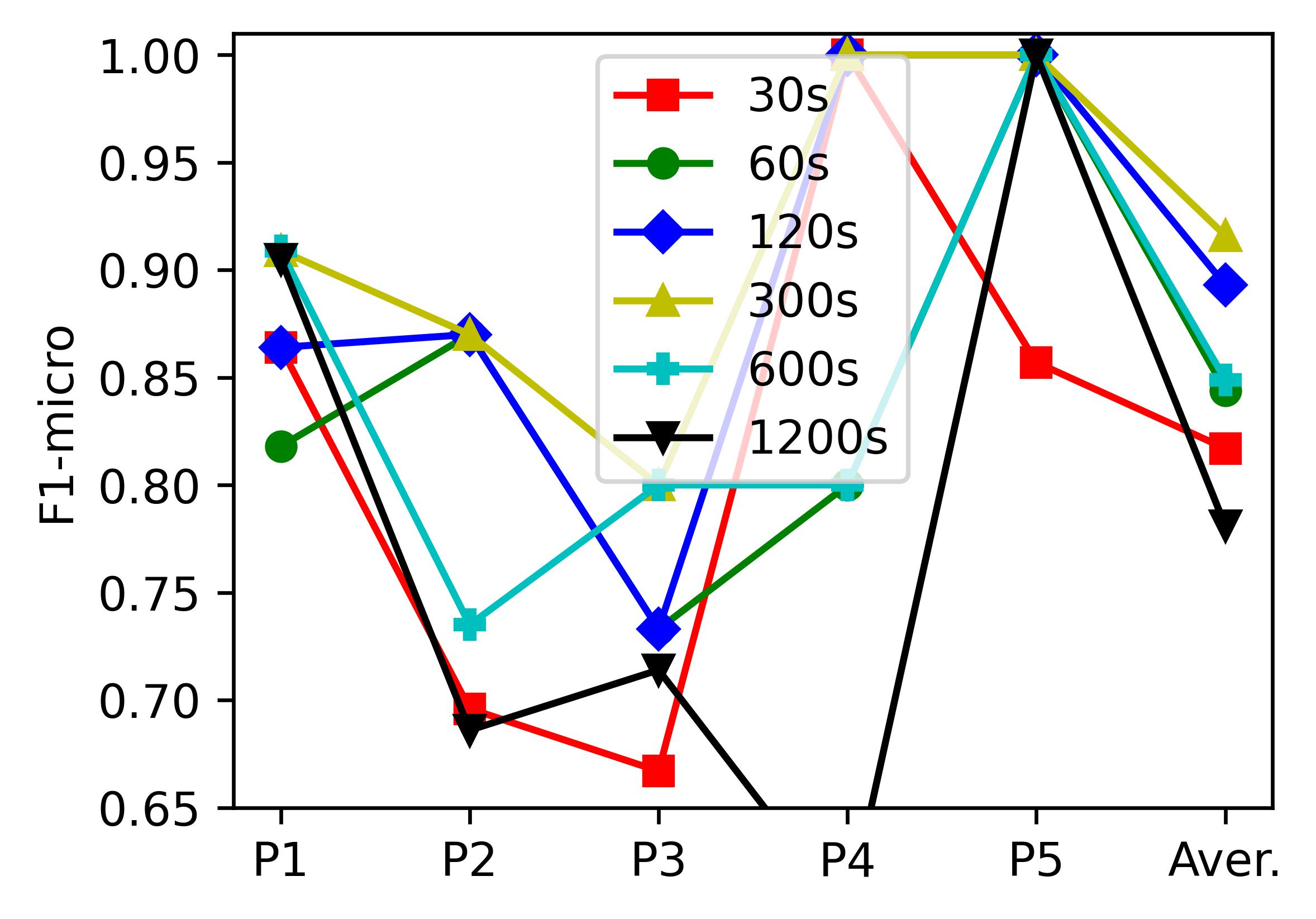}}{Daily-level}
    \caption{F1 score of classifying health states using the Bayesian Network across different temporal windows (30 to 1200 seconds) and aggregated for different time spans (record-level, 8-hour level, and daily-level) for five participants (P1--P5).}  
    \label{fig:timescale}
\end{figure}

Fig. \ref{fig:timescale} provides a comprehensive overview of the Bayesian Network's performance at different time scales, both for training and testing across all activities. Employing different temporal windows from 30s to 1200s affects the performance of the model for health state inference. Notably, in our experiments, the 300s temporal window exhibited the best performance, achieving the highest accuracy across all five participants (P1--P5) when distinguishing between normal and weakness health states across all activity types. For three participants (P1, P3, and P5), the 600s window performed equally well. 

\begin{figure}[h]
    \centering
    \includegraphics[width=0.39\linewidth]{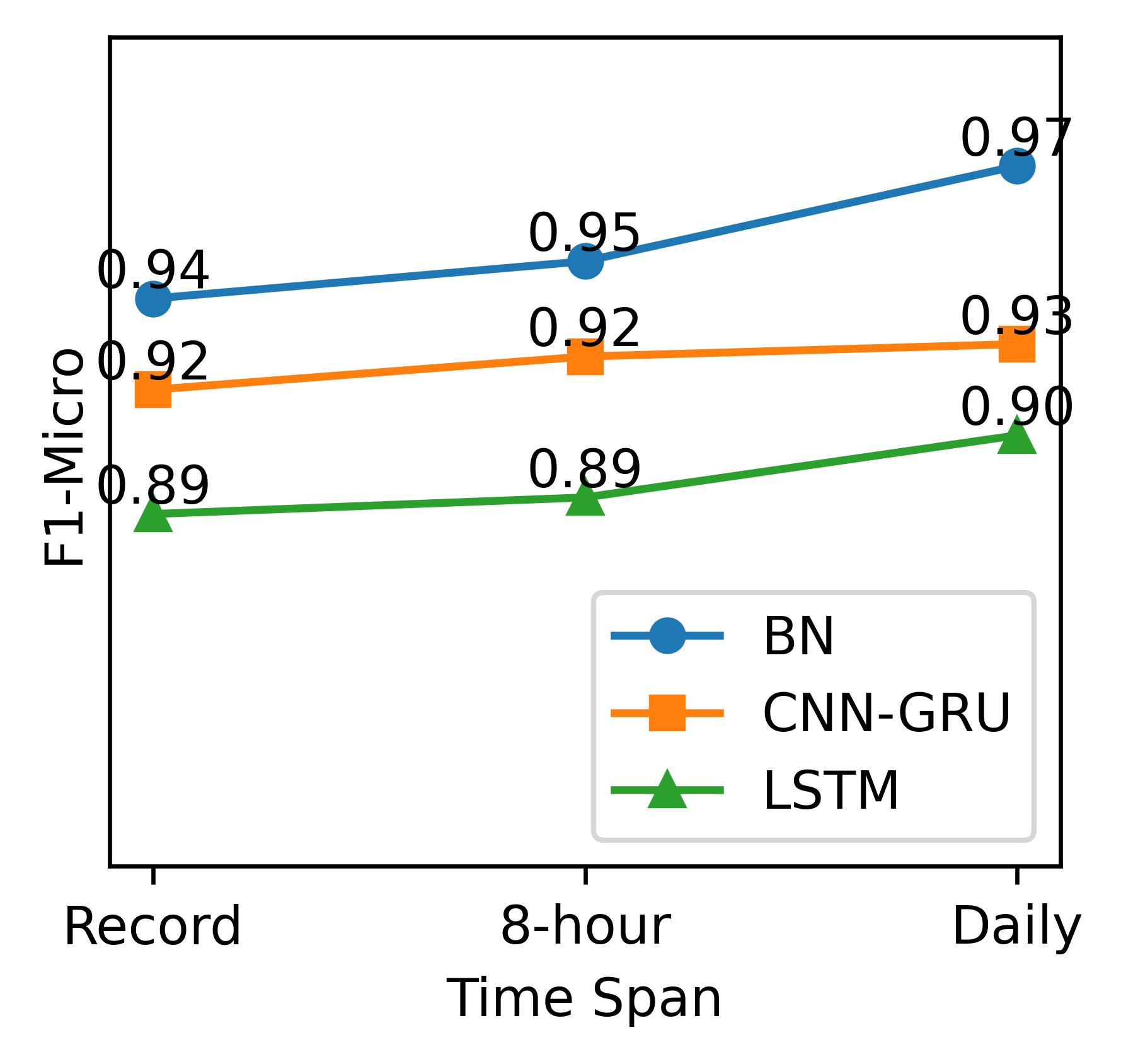}
    \caption{Comparison of the optimal performance of three models (Bayesian Network, CNN-GRU, and LSTM) for health state classification on our dataset, at record-level,  8-hour level, and daily-level, following activity selection and temporal window selection.}
    \label{fig:bestperformance}
\end{figure}

Moreover, further refinement of the model by selecting the best activity combinations alongside the optimal temporal window of 300s significantly improved the accuracy. It reached 0.95 ($\sigma=$ 0.07) at the 8-hour level and 0.97 ($\sigma=$ 0.04) at the daily-level, averaged for all participants. For comparison, we employed two deep convolutional neural networks, CNN-GRU and LSTM, to implicitly learn features from the data without feature engineering. 
Our results indicate that BN outperformed the deep models on our dataset with hand-crafted features (see Fig. \ref{fig:bestperformance}), when considering both optimal time scale and activity combinations. This highlights the robust performance of the BN model and its ability to interpret complex relationships between variables, making it a good choice for modeling features, activities, and health conditions in this context.

\subsection{Feature and Activity Ranking Results} \label{sec:reA}
\begin{figure}
    \centering
    \footnotesize
    \stackunder[5pt]{\includegraphics[width=0.19\linewidth]{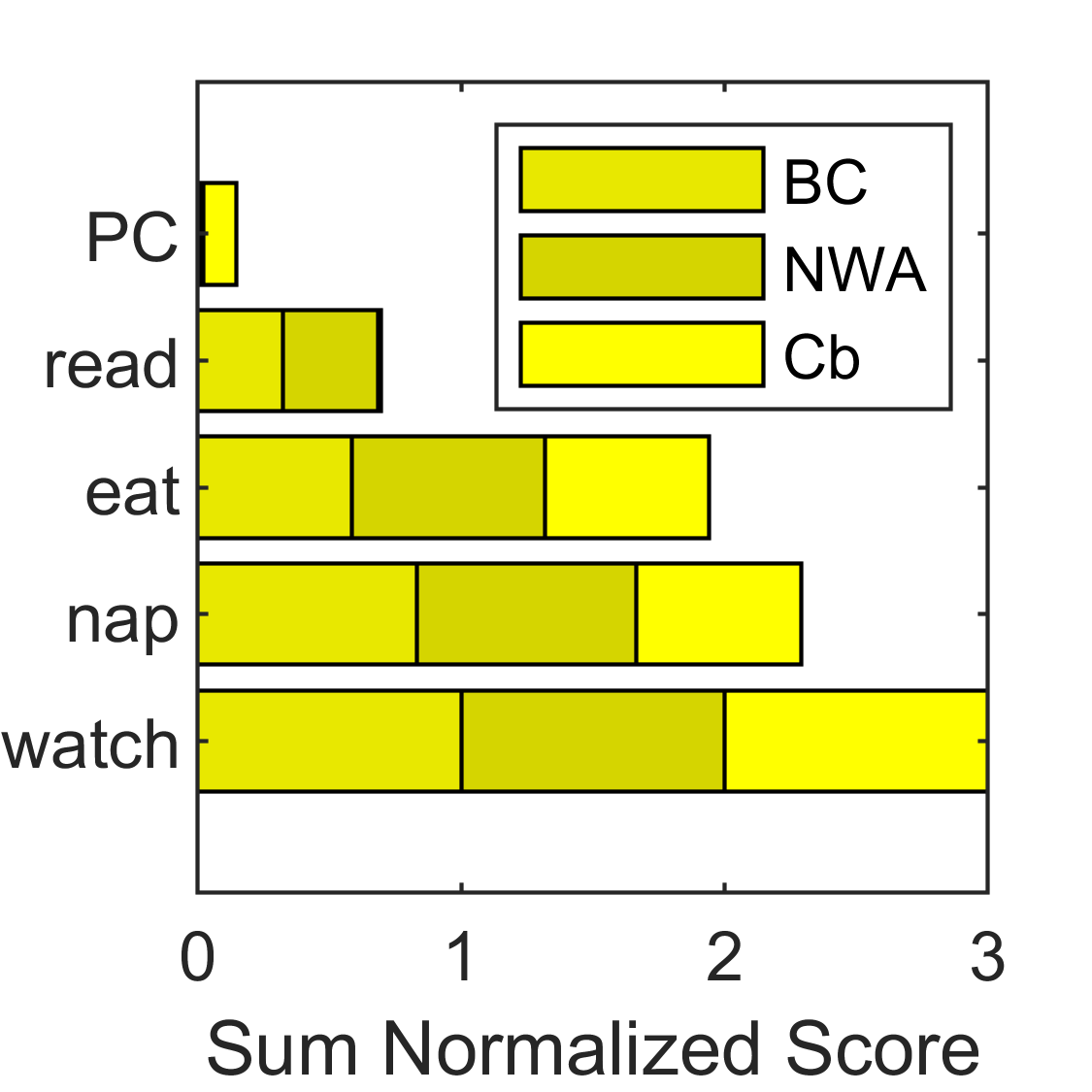}}{P1}
    \stackunder[5pt]{\includegraphics[width=0.19\linewidth]{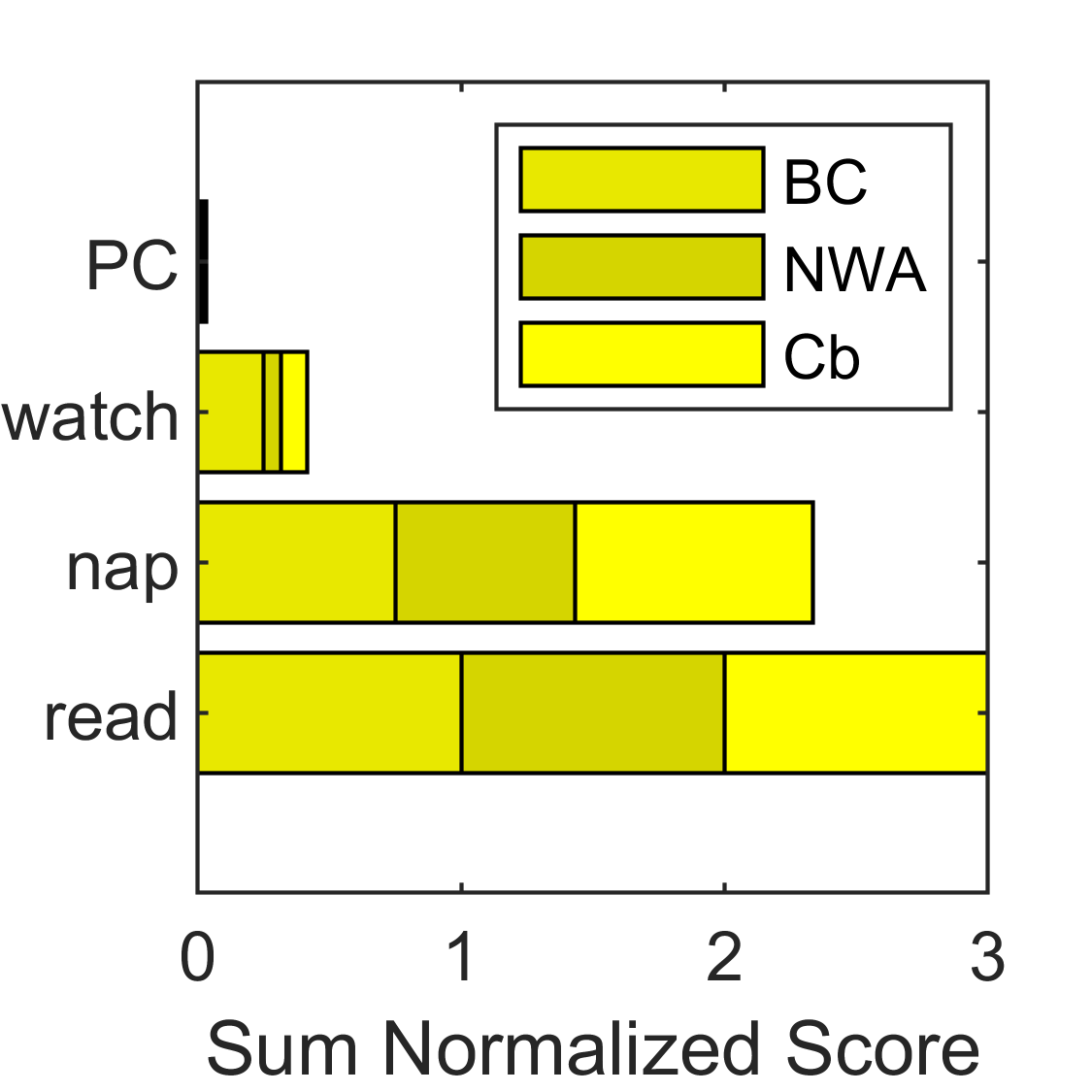}}{P2}
    \stackunder[5pt]{\includegraphics[width=0.19\linewidth]{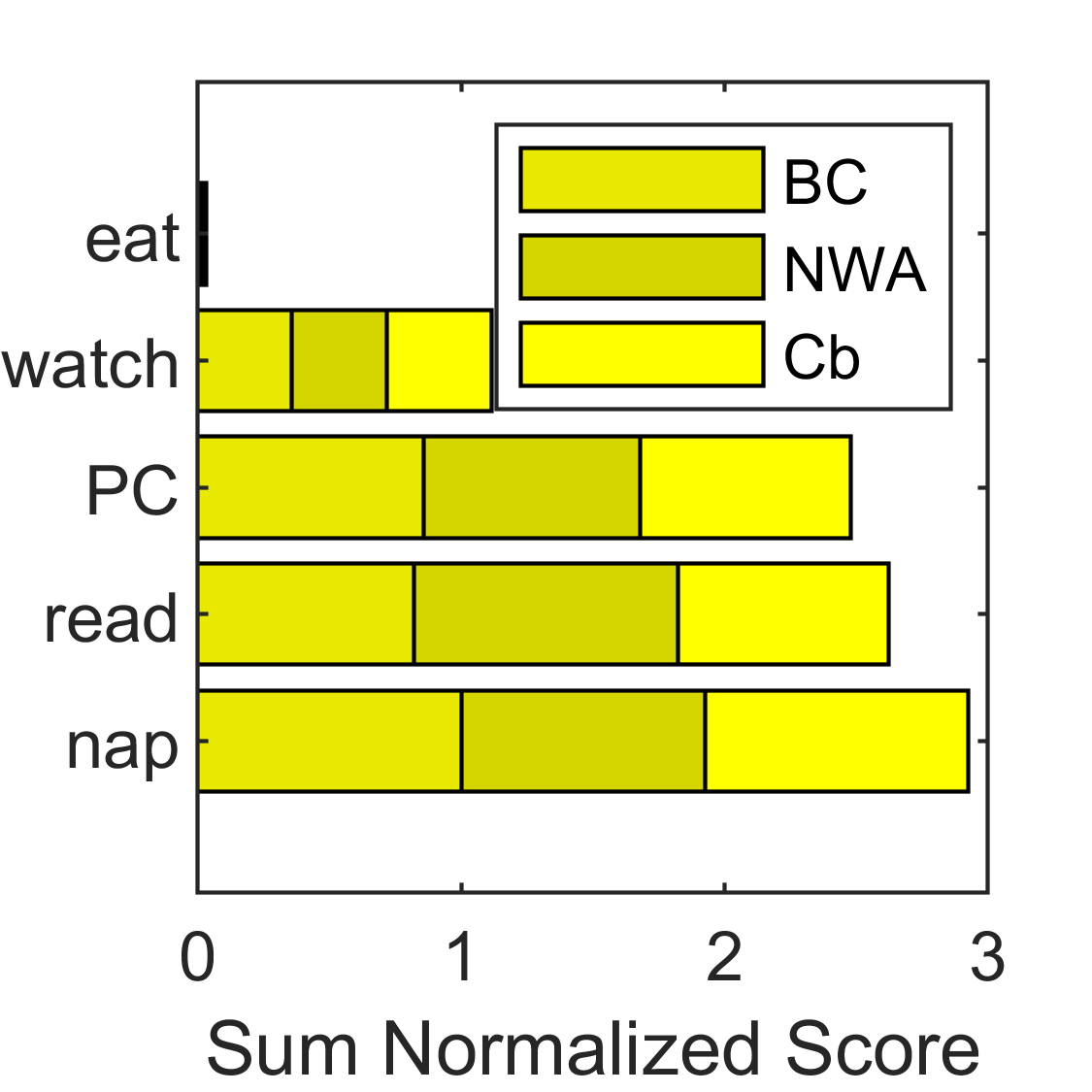}}{P3}
    \stackunder[5pt]{\includegraphics[width=0.19\linewidth]{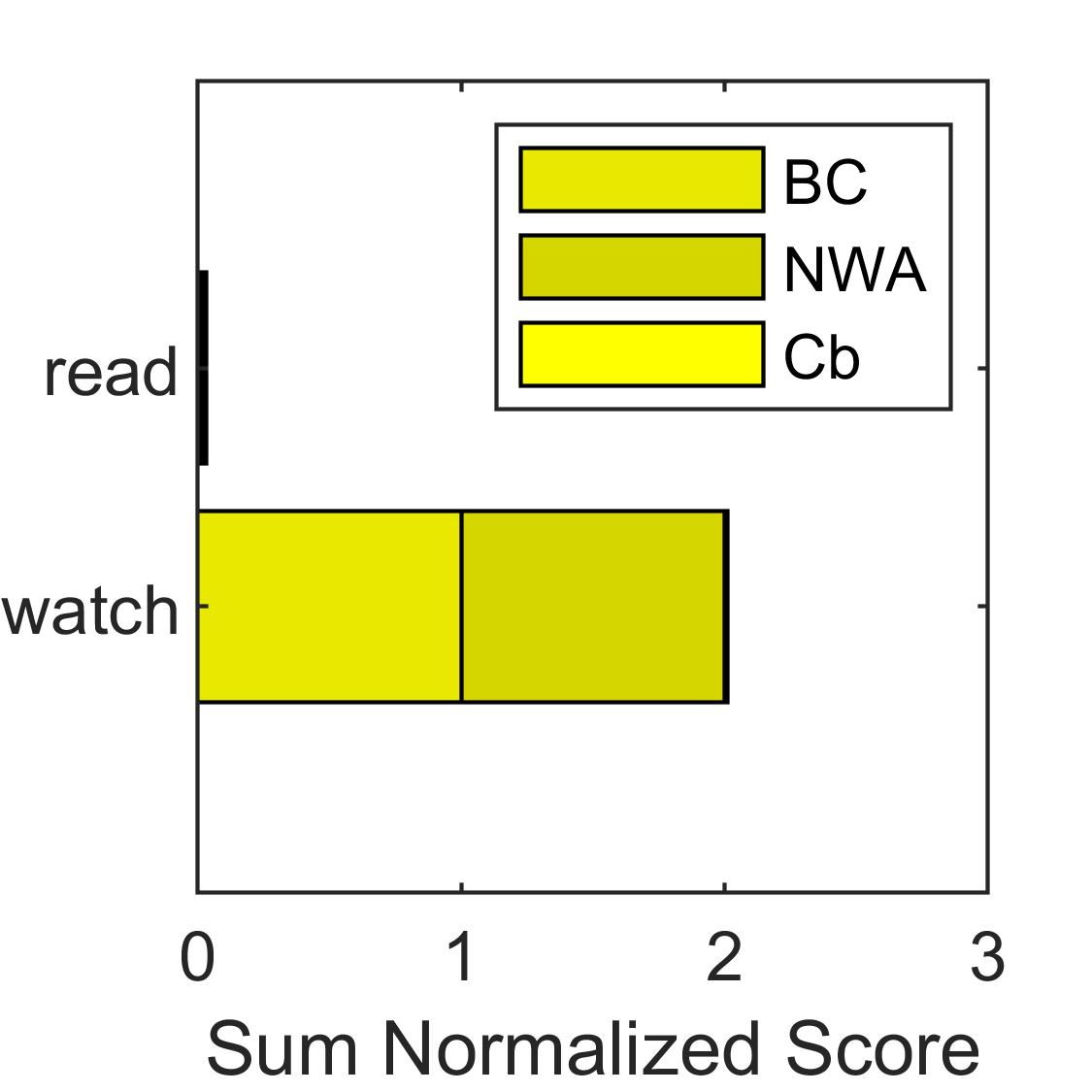}}{P4}
    \stackunder[5pt]{\includegraphics[width=0.19\linewidth]{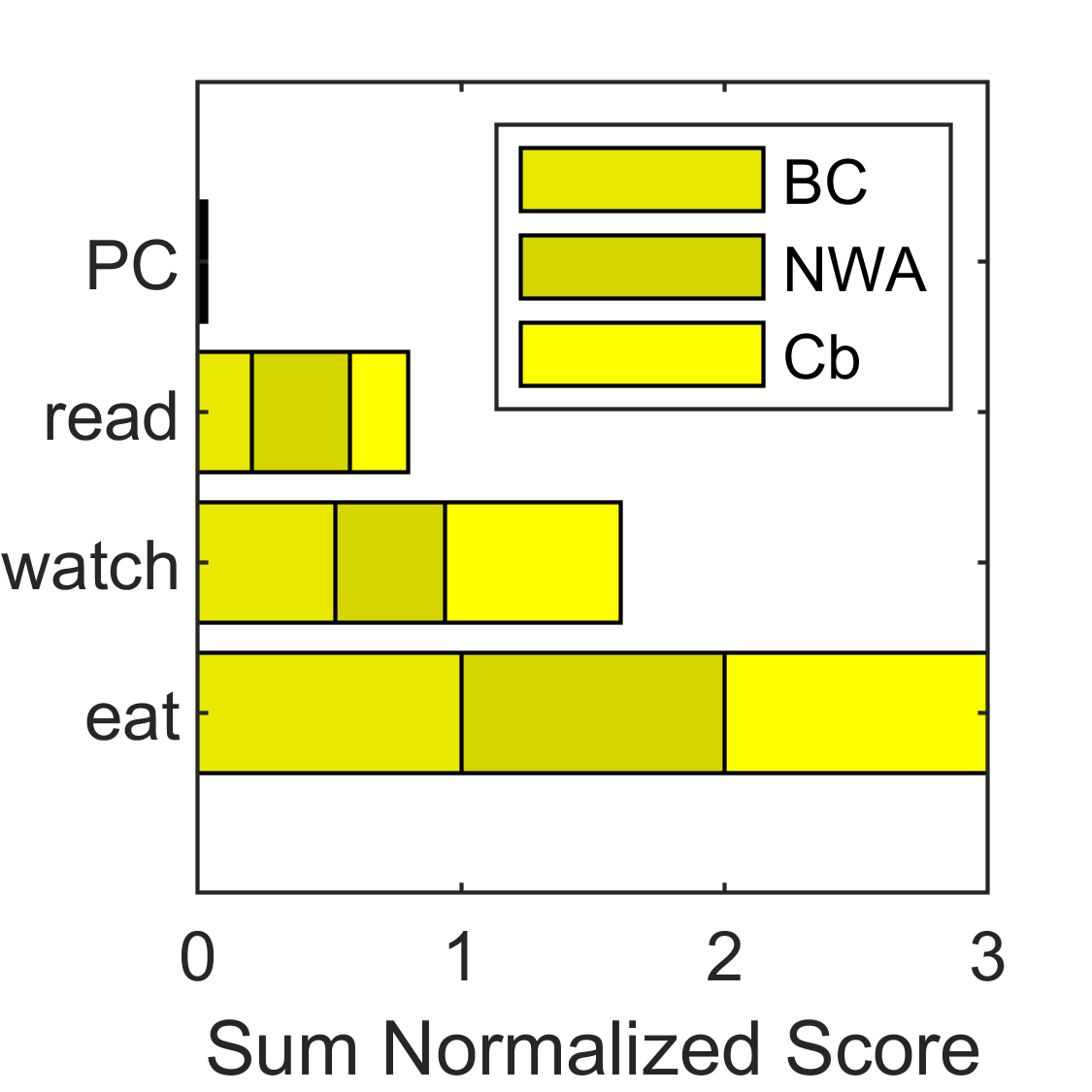}}{P5}
    \caption{Activity ranking results for each participant, after aggregation with Borda Count (BC), Normalization and Weighted Average (NWA), and Consensus-based (Cb). {The horizontal axis displays the final ranking scores by Equation \ref{eq:aggscore}. The vertical axis lists each activity, arranged from top to bottom in ascending order of their ranking scores}, {where larger scores are better.} {The results indicate that "watch" and "nap" ranked among the top two most important activities across participants. Conversely, "PC" was consistently ranked as the least important activity.}
    } 
    \label{fig:aheach}
\end{figure}

The results of the activity ranking can be found in Fig. \ref{fig:aheach}.
The result shows activity `nap' appeared three times in the top 2 
rankings (i.e., highest ranking scores among activities), and `watch' also appeared three times. 
On the contrary, `PC' was consistently ranked as the least important activity, appearing three times as such. 
Two participants (P1 and P3) participated in all five types of activities, while the others were not available for some of the activities.

\begin{figure}
    \centering
    \footnotesize
    \stackunder[5pt]{\includegraphics[width=0.19\linewidth]{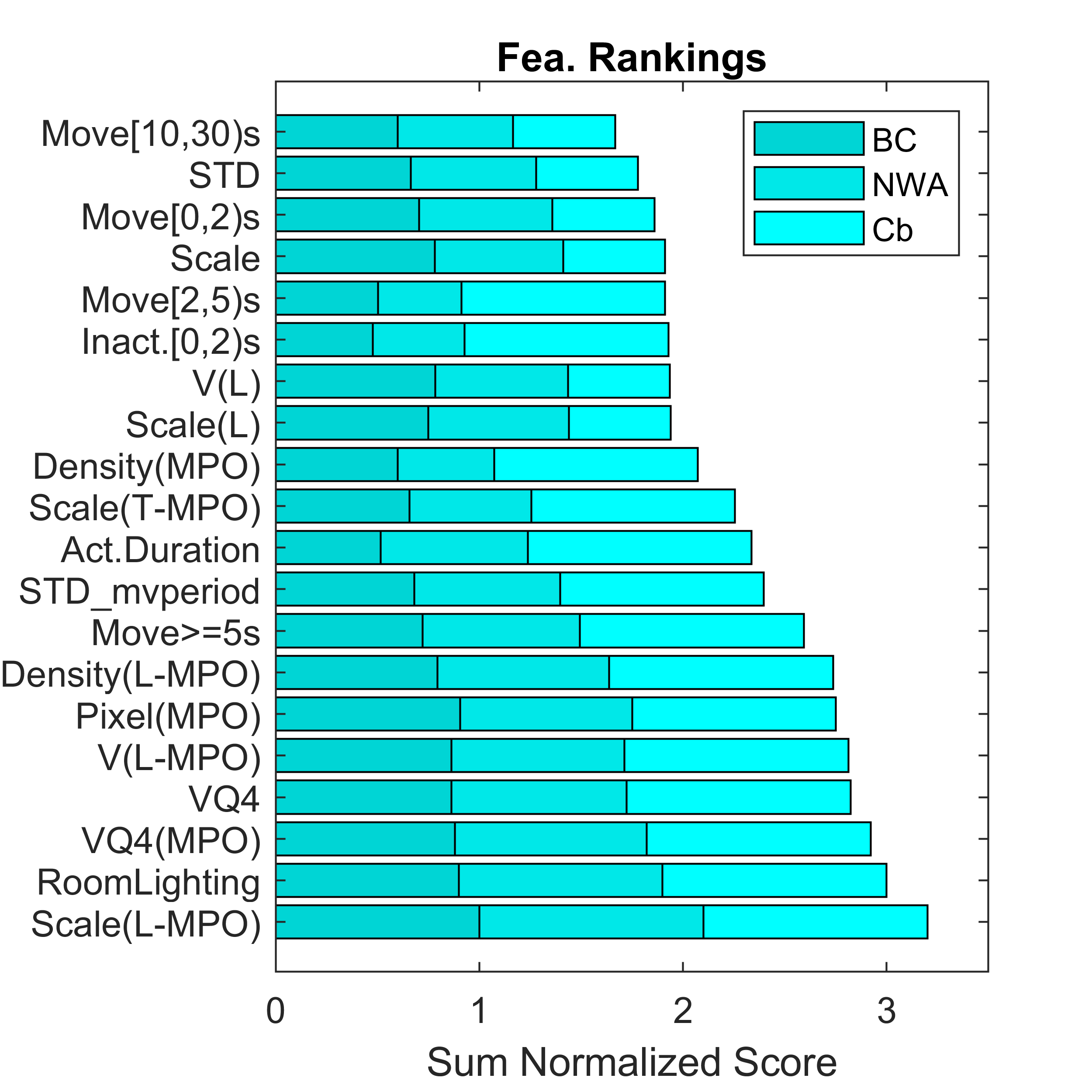}}{P1}
    \stackunder[5pt]{\includegraphics[width=0.19\linewidth]{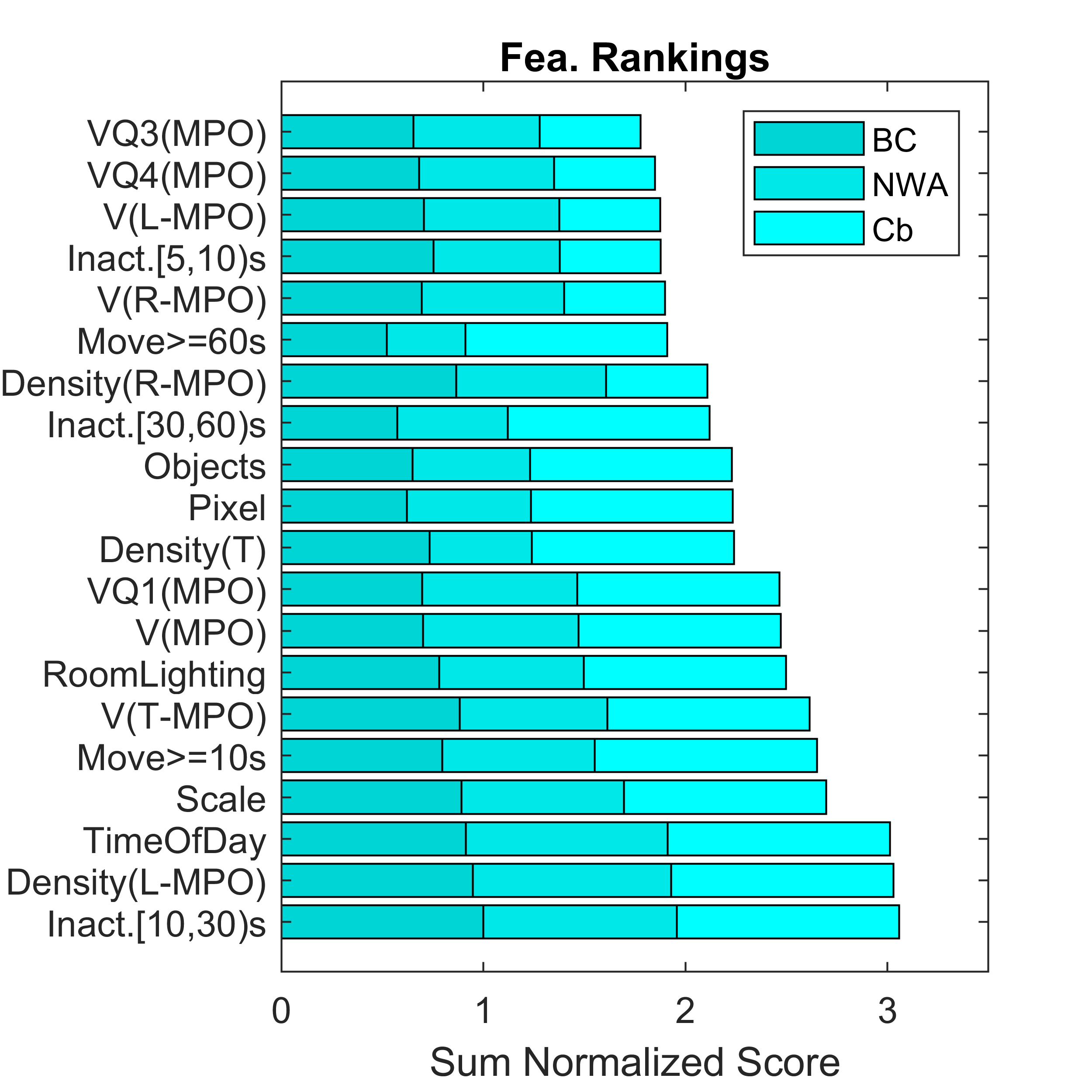}}{P2}
    \stackunder[5pt]{\includegraphics[width=0.19\linewidth]{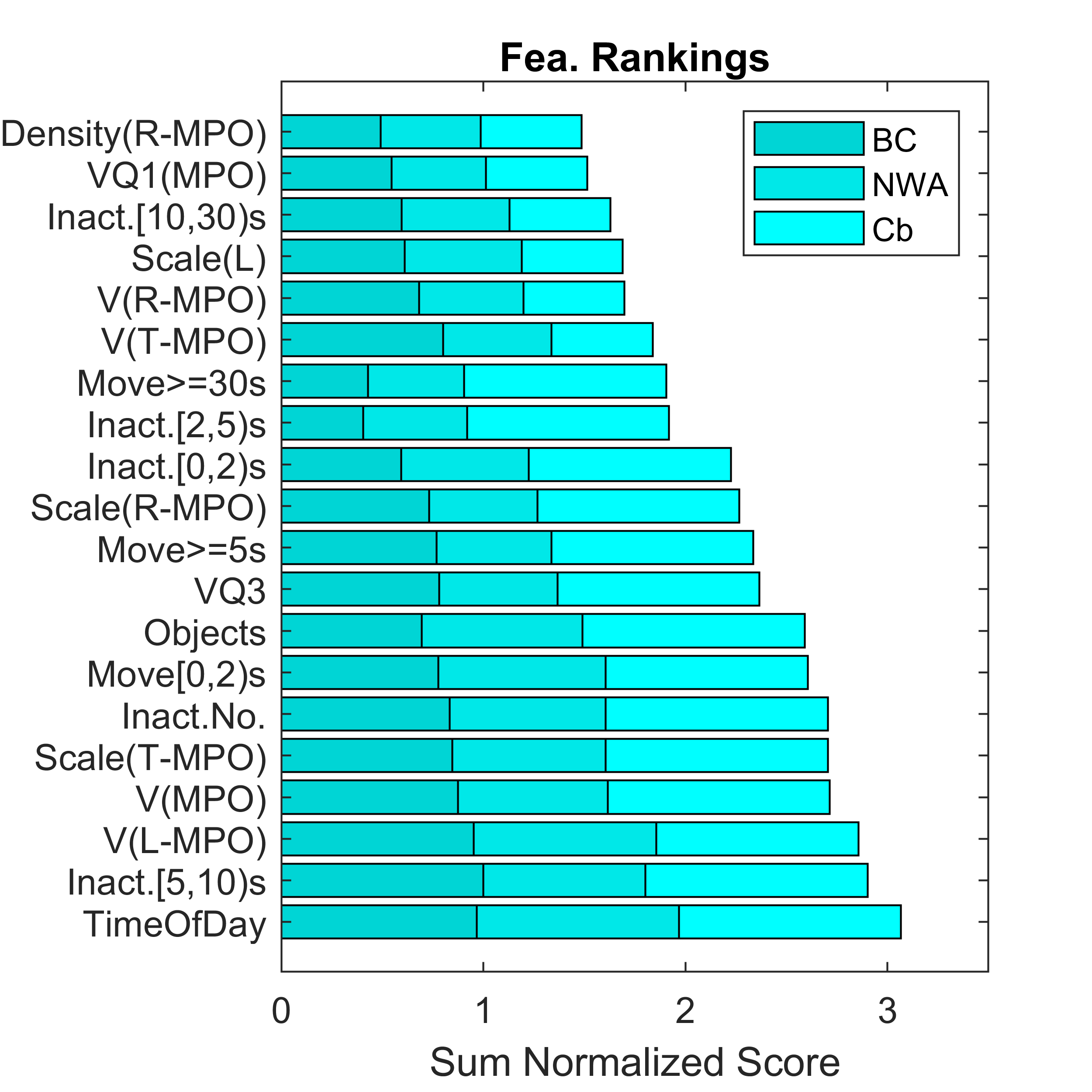}}{P3}
    \stackunder[5pt]{\includegraphics[width=0.19\linewidth]{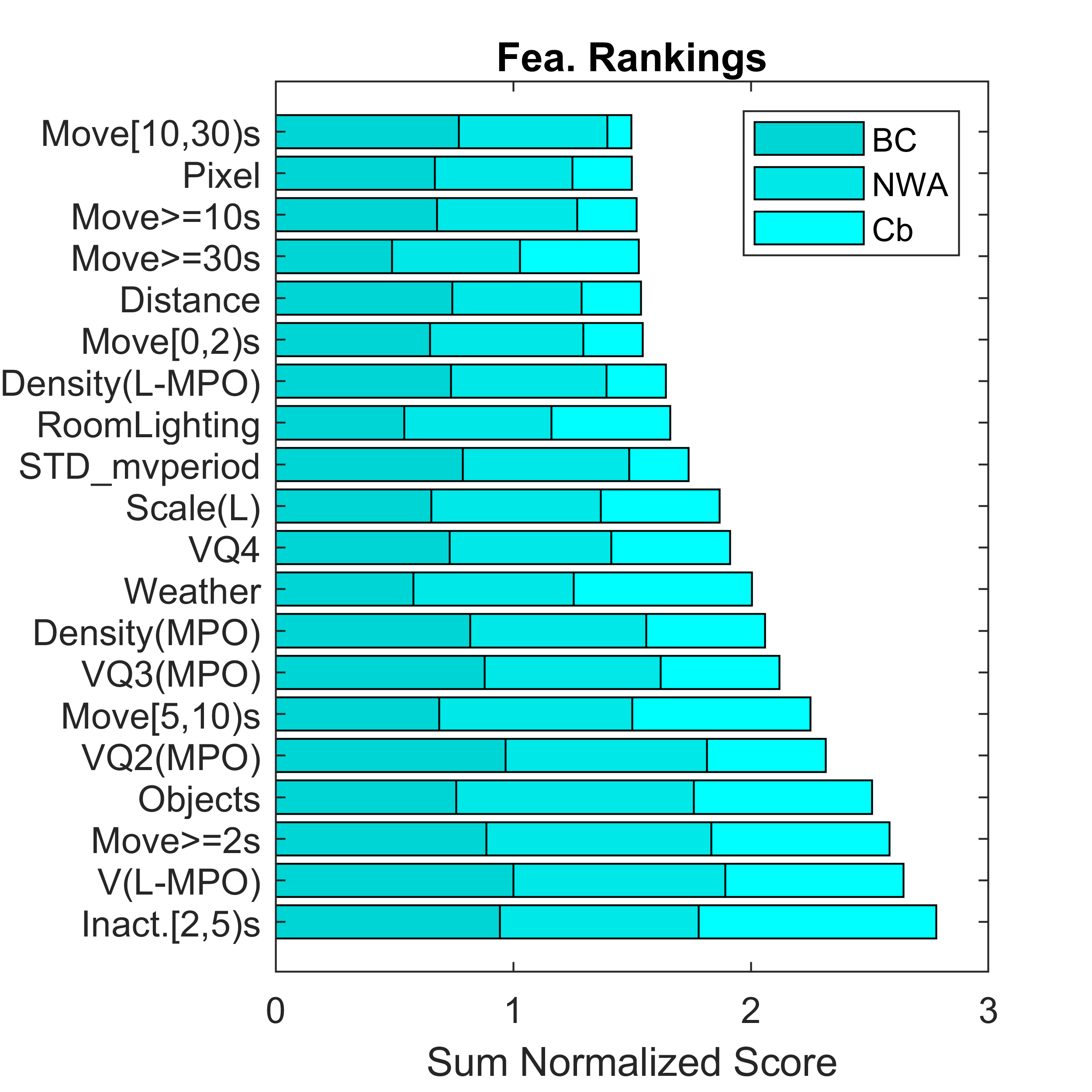}}{P4}
    \stackunder[5pt]{\includegraphics[width=0.19\linewidth]{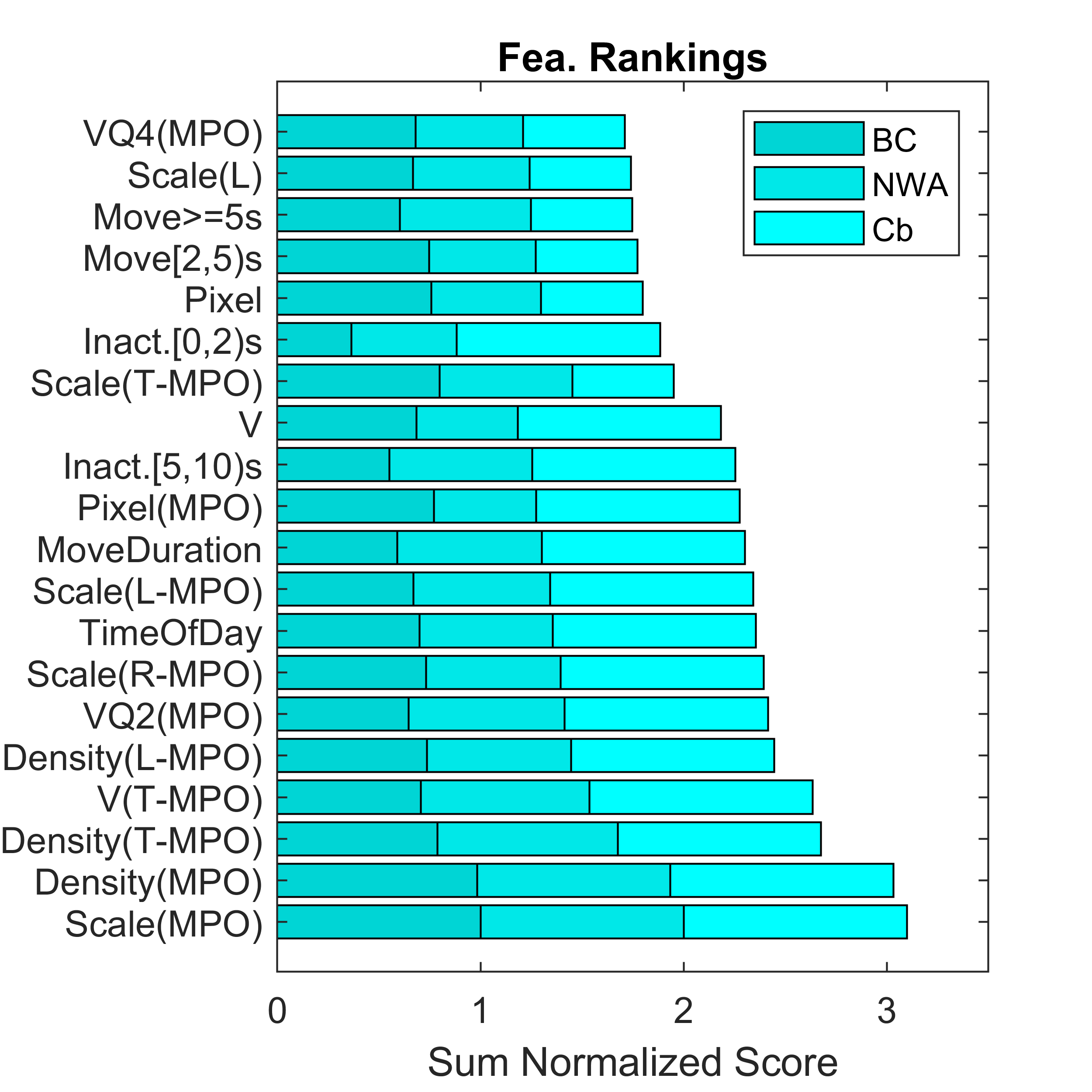}}{P5}
    \caption{Feature ranking results for each participant. 
    Aggregated with Borda Count (BC), Normalization and Weighted Average (NWA), and Consensus-based (Cb). {The horizontal axis displays the final ranking scores by Equation \ref{eq:aggscore}, {where larger scores are better.} The vertical axis lists top-20 features for each participant, arranged from top to bottom in increasing order of their final ranking scores.}
   {Fine-grained motion features (scale, density, and speed) and inactivity distributions emerged as significant indicators of health states, with notable variations in the top features across participants.}
    }
    \label{fig:fheach}
\end{figure}

The top-20 ranked features for each participant are displayed in Fig. \ref{fig:fheach}, while the overall feature ranking for all participants is presented in Table \ref{tab:topfeatureseachp}. The results highlight the significance of both movement and inactivity features. Among the top 10 features averaged for all participants, movement speed appeared three times, movement density also occurred three times, and movement scale appeared four times. Additionally, among the top 20 features, inactivity distribution was observed three times.

When comparing weakness states to normal states within each individual, specific changes were observed for each participant based on their own feature rankings. 
Table \ref{tab:topfeatureseachp} illustrates each participant's percentage change in the top-ranked feature groups, showing the average percentage change (\%) in feature values from the normal to the weakness state. A negative change in movement speed or scale, or a positive change in inactivity-related features, may indicate a deterioration.
{The results so far suggest that there are no features that are generally useful across all participants. See Section \ref{sec:discuss} for more discussion.}

For Participant 1, the {feature} rankings suggested a reduction in the movement scale on the non-dominant side of the body and a decrease in peak movement speed when comparing weakness to normal states. More specifically, the movement scale and speed in the left side of the body exhibited reductions of -13.1\% and -14.9\%, respectively, across all activities.
{That is, they moved the left side of their body less distance and slower.}

Participant 2's {feature} rankings indicated a reduction in short to middle inactivity periods but an increase in long inactivity periods, with a decrease of -31.9\% in the 0- to 2-second range and a large increase of +196.1\% in the greater-than-or-equal-to-60-second range.
{In other words, they had fewer short rests, but many more longer rests.}

Participant 3's feature rankings suggested paying attention to the movement speed during the movement period only, which decreased by -18.1\% (and -7.4\% on the left side of the body). Similar to Participant 2, there were evident changes in inactivity distribution.

For Participant 4, the feature rankings suggested focusing on the movement speed during the movement period only, revealing increases for the second and third quartiles, and on the left side of the body. Movement distribution also changed, suggesting an increase in middle-duration movement and a reduction in long-duration movement, relative to an increase in long-duration inactivity.

Finally, for Participant 5, the feature rankings highlighted reductions in movement scale and density during the movement period only. Notably, the movement scale of the left side of the body decreased by -20.5\%, and the speed of the upper-body part decreased by -5.4\%. Furthermore, the first to fourth quartiles of movement speed in the movement period only exhibited obvious decreases.

\subsection{Anomaly Detection using a Bayes Net}
{
The preceding sections investigated the distinction between normal and weak health states by framing it as a classification problem with simulated weakness data. We then demonstrated the model's ability to differentiate between these states and identified promising indicators for detecting such changes. However, real-world health data often presents challenges: controlling health states is difficult, and imbalanced data is common, with fewer abnormal cases than normal ones. This makes traditional classification approaches less practical.}

{
Anomaly detection, where a model is trained only on normal data is used to identify abnormal situations, offers a more suitable solution in these scenarios. We can leverage it to detect rare and imbalanced abnormality instances.
Therefore,  a Bayesian network using only data from normal health states was constructed. 
This ``normal model'' aims to accurately capture the typical patterns within normal healthy days. 
Its ability to identify abnormal cases (weak days) based on their deviations from the expected normal patterns was then tested.}

Specifically, we grouped monitoring records at the daily-level and trained a Bayes Net model for each participant solely on samples from normal days. {The net can be seen in Fig. \ref{fig:BN}.}
The model construction can be represented as:
\begin{equation}
 \mathcal{M}_{nor} =BN(\mathcal{D}_{nor})
\end{equation}
where $\mathcal{D}_{nor} = \{\mathcal{(H} = h_{nor}, \mathcal{A}, \mathcal{F})\}$.
Here, $\mathcal{M}_{nor}$ denotes the normal model, $\mathcal{D}_{nor}$ is the data with normal health states $h_{nor}$.

{
A leave-one (day)-out strategy was used for training and testing, i.e., during each training iteration, samples from one normal day were excluded for testing purposes.  
For training, only normal days with multiple monitoring records were used, excluding those with only one sample, to obtain a more stable model with less variation.
After training,  the model was evaluated using data from weak days and averaging the output log-likelihoods for each day.}

\begin{figure}[h]
    \centering
    \includegraphics[width=1\linewidth]{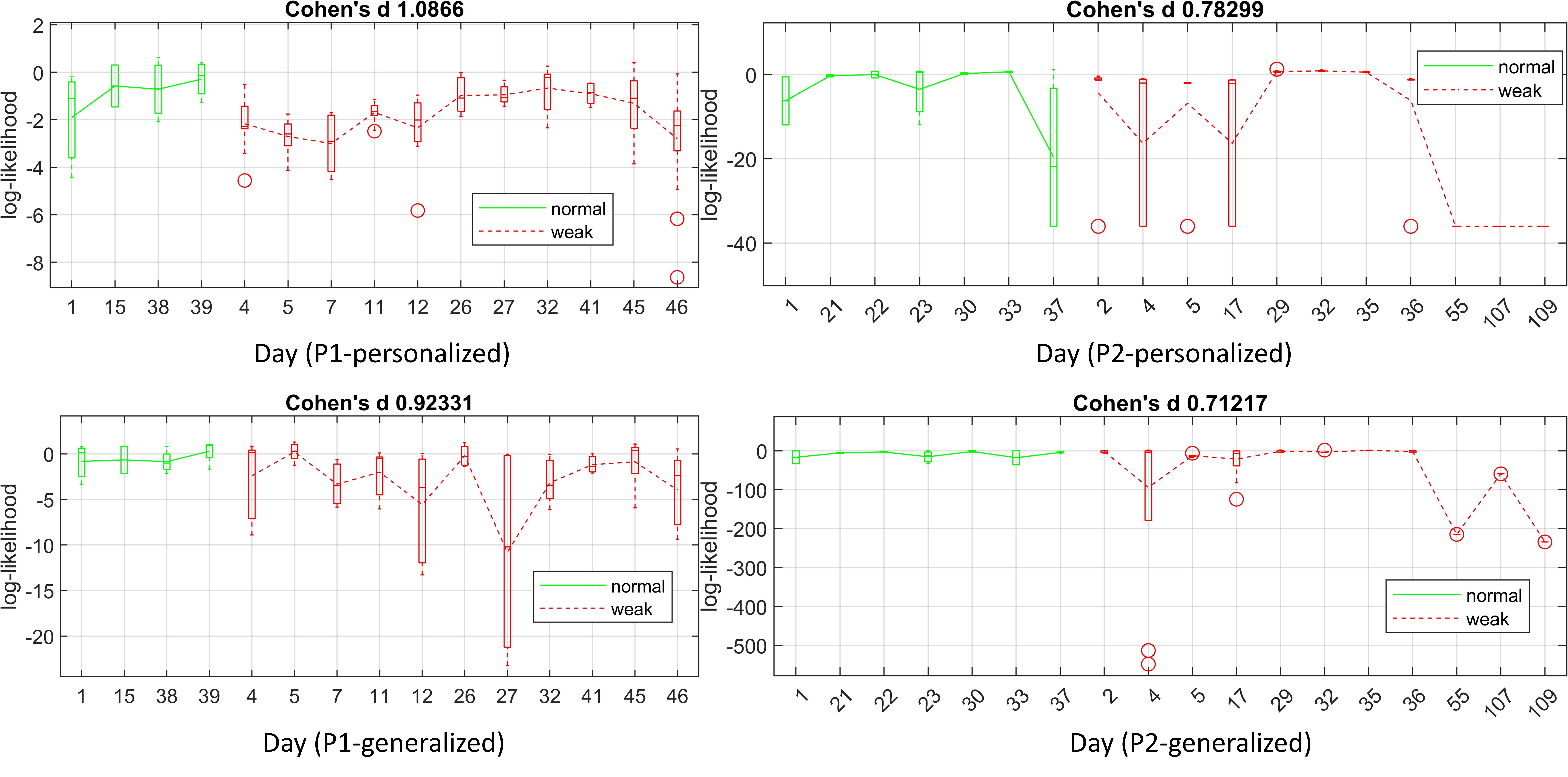}
    \caption{Output log-likelihoods generated by the ``normal models'' for the normal and weak days of two participants.     
    The horizontal axis displays the monitoring day index for each participant (P1 or P2). Normal days with zero or one record were excluded from model training. The solid line depicts the mean log-likelihood for normal days, while the dotted line represents the mean for weak days.  
    {The top row displays the output of models built with personalized best features (Top-2 and Top-3 features for P1-P2, respectively). The bottom row shows the output of models built with top-5 generalized features. Both rows demonstrate noticeable and meaningful differences between the means of the normal and weakness groups.}
    }

    \label{fig:normalModels}
\end{figure}

Fig.\ref{fig:normalModels} shows the test sample output log-likelihoods for normal and weak days for two participants (participants with sufficient data for individualized model building). 
{
For personalized models, we first select within the top five ranked features for each participant from the list presented in Table \ref{tab:topfeatureseachp} top. Then, to assess feature generalizability, we built normal models using the top five features ranked across all participants, as detailed in the bottom section of Table \ref{tab:topfeatureseachp}.}
As visualized, the normal models consistently assign higher log-likelihoods to the normal test sample days compared to weak days.
Furthermore, Cohen's d between the log-likelihood values of the two groups was calculated. The results of over 0.9 and 0.7 indicate large and moderate effect sizes, respectively, signifying noticeable and meaningful differences between the means of the normal and weak groups.
Moreover, using personalized feature sets, when supported by sufficient personal data, can improve the detection of abnormal health states compared to generic feature sets.


\section{Discussion\label{sec:discuss}}
The focus of this study was on observing natural and common daily scenarios and identifying health state changes caused by weaknesses (simulated by an exercise workout session). 
The objective was to gain insights into the effectiveness of automatically quantifying behavioral changes and explaining these changes. We specifically targeted common daily activities that people frequently engage in at home, such as reading, napping, watching TV, using a PC, and eating, particularly while sitting on their favorite chairs or couches. This scenario setting provided a simple and feasible solution for monitoring their behaviors unobtrusively over the long term. 

We utilized a fixed RGB-D camera with a specific emphasis on capturing upper-body movements. During real-time, anonymous processing, we extracted explicit motion, inactivity, and environmental features while considering their dependencies. The results showed that our method could distinguish between normal and weak states effectively. By selecting the appropriate activity and suitable temporal window for behavioral features and further aggregating data over longer time spans, our method achieved an accuracy of 0.97 {($\sigma=$ 0.04)} at the daily-level with 5-fold cross-validation.

When performing long-term monitoring, including all kinds of activities during a monitoring period is not always the best choice. 
{The average F1-micro scores shown in Fig. \ref{fig:acc}, when averaged across all classifiers, participants, and activities, is 0.79.} 
This might be due to significant variations in feature values among different activities, which can be larger than the differences between health states, as illustrated in Fig. \ref{fig:example}.
Selecting prominent activities is beneficial for detecting a change in health states as the contribution of each activity is uneven, with participants exhibiting distinct behavioral patterns in each activity. After selecting the optimal combination of activities, the  {F1-score}
 improved by about 10\%. 
Here, we chose combinations of activities that accounted for more than 40\% of the data coverage but did not select the top-1 activity. This decision was made because some activities are naturally rare, which could introduce bias in classification results due to the limited number of samples. 
Not knowing the activity types slightly reduces the performance of health state classification; however, it may help alleviate the difficulty of automatic activity recognition, especially in cases where manually recording the activity type is challenging. When the activity type is unknown, we achieved an average {F1} of 0.819 on the record-level using RF and SVM, representing a 1.4\% drop compared to training with known activity types. (BN is excluded here due to model complexity, as the sample size in each activity on the record-level is too small, resulting in overfitting for Participant 5).

Selecting suitable temporal windows also enhances the performance of classifying health states. 
The results in Fig. \ref{fig:timescale} show that 5 to 10-minute windows are the best segments for feature extraction using BN. This approach is feasible in terms of computational cost and data storage for long-term monitoring, allowing the method to record statistics at regular intervals, such as the average movement speed or motion and inactivity distributions. Furthermore, aggregating the results from temporal windows into longer time spans effectively enhances performance. When there are more monitoring records within a specific period, such as within 8 hours or a day, the results become more robust. 
For example, Participant 1 has more than 8 activity monitoring records per day, and aggregation can enhance 
{F1} 
by more than 15\% when moving from the record-level to the daily-level. 
On average,  a 5\% improvement across all participants was observed. 
In conclusion, extracting behavioral features every several minutes and then aggregating them into longer time spans is a sound approach for making decisions when inferring health states.

\begin{table}[]
\footnotesize
\caption{ 
{(Top) Top-5 behavioral features ranked for each participant (from Fig. \ref{fig:fheach}), showing their {average}
percentage change (\%) from normal to weakness.} 
{
(Bottom) Top-10 behavioral features ranked across all participants, determined by averaging the ranking scores of each feature among all participants.}
}
\begin{tabular}{lllll}
\toprule
P1 (R-handed)   & P2 (R-handed)    & P3 (R-handed)     & P4 (R-handed)     & P5 (R-handed)   \\
\midrule
\begin{tabular}[c]{@{}l@{}}
{Scale(L-MPO)} -13.1\\ {VQ4(MPO)} -9.3\\ 
{VQ4} -17.3\\ {V(L-MPO)} -14.9 \\  {Pixel(MPO)}  -18.7\\ \end{tabular} & 
\begin{tabular}[c]{@{}l@{}}  {Inact.10-30s} -7.9\\ 
{Density(L-MPO)} +21.2\\ {Scale} -94.2 \\ {Move.geq10s} -16.1 \\ 
{V(T-MPO)}	+29.9
 \end{tabular} &
\begin{tabular}[c]{@{}l@{}} {Inact.5-10s}  +63.3\\ 
{V(L-MPO)} -7.4\\ {V(MPO)} -18.1\\ {Scale(T-MPO)} -8.6 \\ {Inact.No.}	+22.1
\end{tabular} & 
\begin{tabular}[c]{@{}l@{}} {Inact.2-5s} -27.0\\ 
{V(L-MPO)}  +43.1\\  {Move.geq2s} +3.7 \\
{VQ2(MPO)} +36.8 \\ {Move.5-10s} +61.6 
\end{tabular} & 
\begin{tabular}[c]{@{}l@{}}{Scale(MPO)} -5.7\\   {Density(MPO)}  -14.0\\ {Density(T-MPO)} +19.7  \\ {V(T-MPO)} -5.4\\ {Density(L-MPO)} +28.1 \\ 
\end{tabular} \\
\midrule
\multicolumn{5}{l}{
\begin{tabular}[c]{@{}l@{}}Top 10 behavioral features ranked among all participants \\ V(L-MPO): movement speed in the left side of the body in movement period only\\ Density(L-MPO): movement density in the left side of the body in movement period only\\ Scale(T-MPO): movement scale in top part in movement period only\\ VQ4(MPO):  top25\% fastest speed in movement period only\\ Density(MPO): movement density in movement period only\\ Scale(L): movement scale in the left side of the body\\ Pixel: movement pixel count (similar to density)\\ Scale(L-MPO): movement scale in the left side of the body in movement period only\\ V(T-MPO): speed in top body parts in movement period only\\ Scale(R-MPO): movement scale in right body part in movement period only
\end{tabular}
}   \\
\bottomrule
\end{tabular}
\label{tab:topfeatureseachp}
\end{table}

Feature ranking reveals the behavior patterns that are crucial in distinguishing between normal and weakness health states. Among all the features, movement speed and inactivity emerge as strong indicators of weakness, followed by movement scale and movement density.
Notably, fine-grained behavioral features hold greater significance in this context, particularly those related to MPO (movement period only) and L-MPO (left body region in MPO - the non-dominant side of the 5 participants). 
These features consistently appear in the top rankings for each participant (see Fig. \ref{fig:fheach}). This underscores the importance of focusing on the non-dominant region of the body for participants who are all right-handed in our experiment. The non-dominant region predominantly encompasses non-essential movements that can be omitted when not required, whereas the dominant body parts involve movements necessary for specific tasks and cannot be omitted as easily.
Nonetheless, it's important to acknowledge that each participant exhibits their own unique set of optimal behavioral features and preferred activities. Substantial variations in behavioral characteristics exist not only among different activities but also among different participants. The ranking results reveal that there is no single set of unified behavioral features applicable to all participants. 

\begin{figure}[h]
    \centering
    \footnotesize
    \stackunder[5pt]{\includegraphics[width=0.39\linewidth]{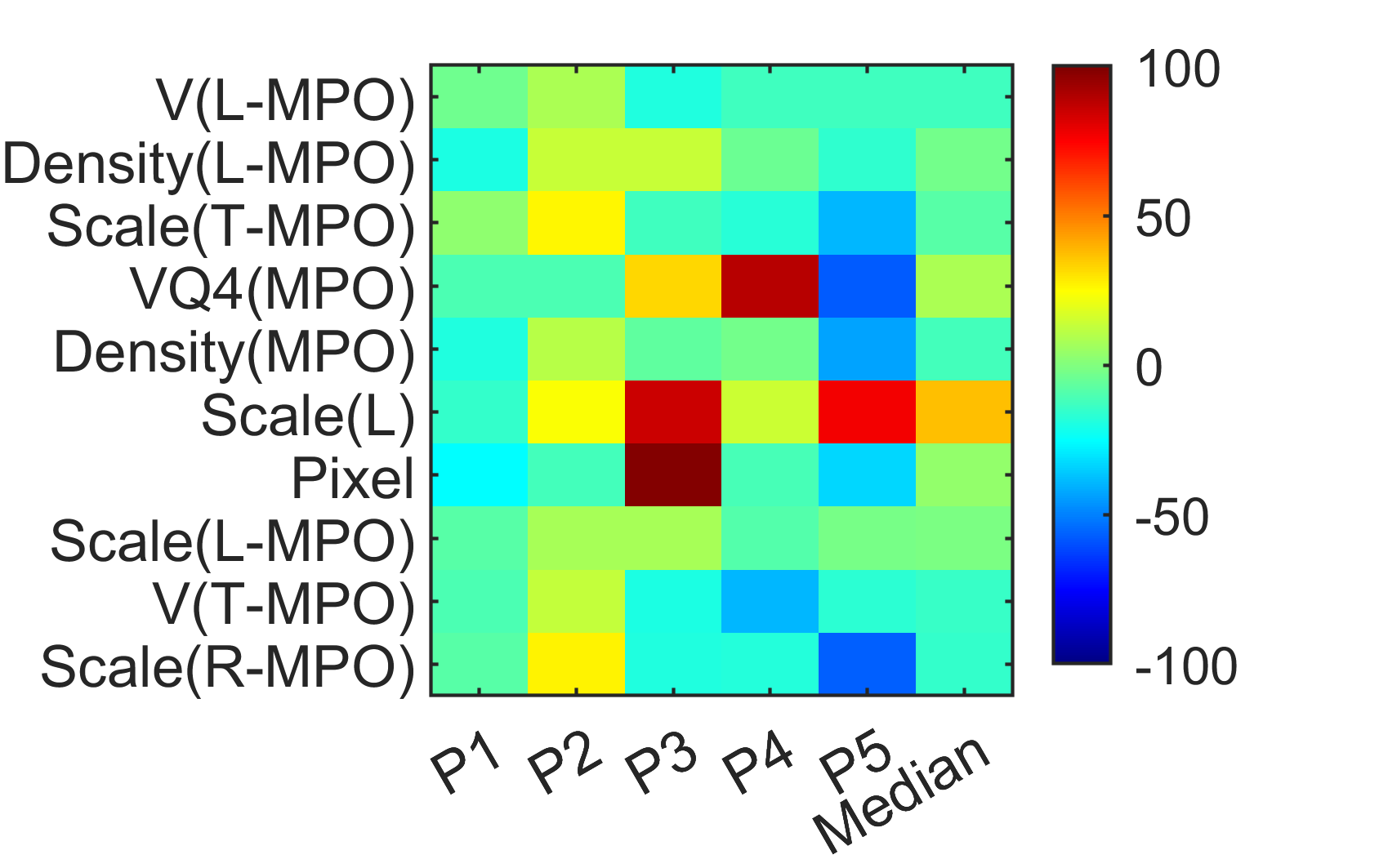}}{Watch}
    \stackunder[5pt]{\includegraphics[width=0.39\linewidth]{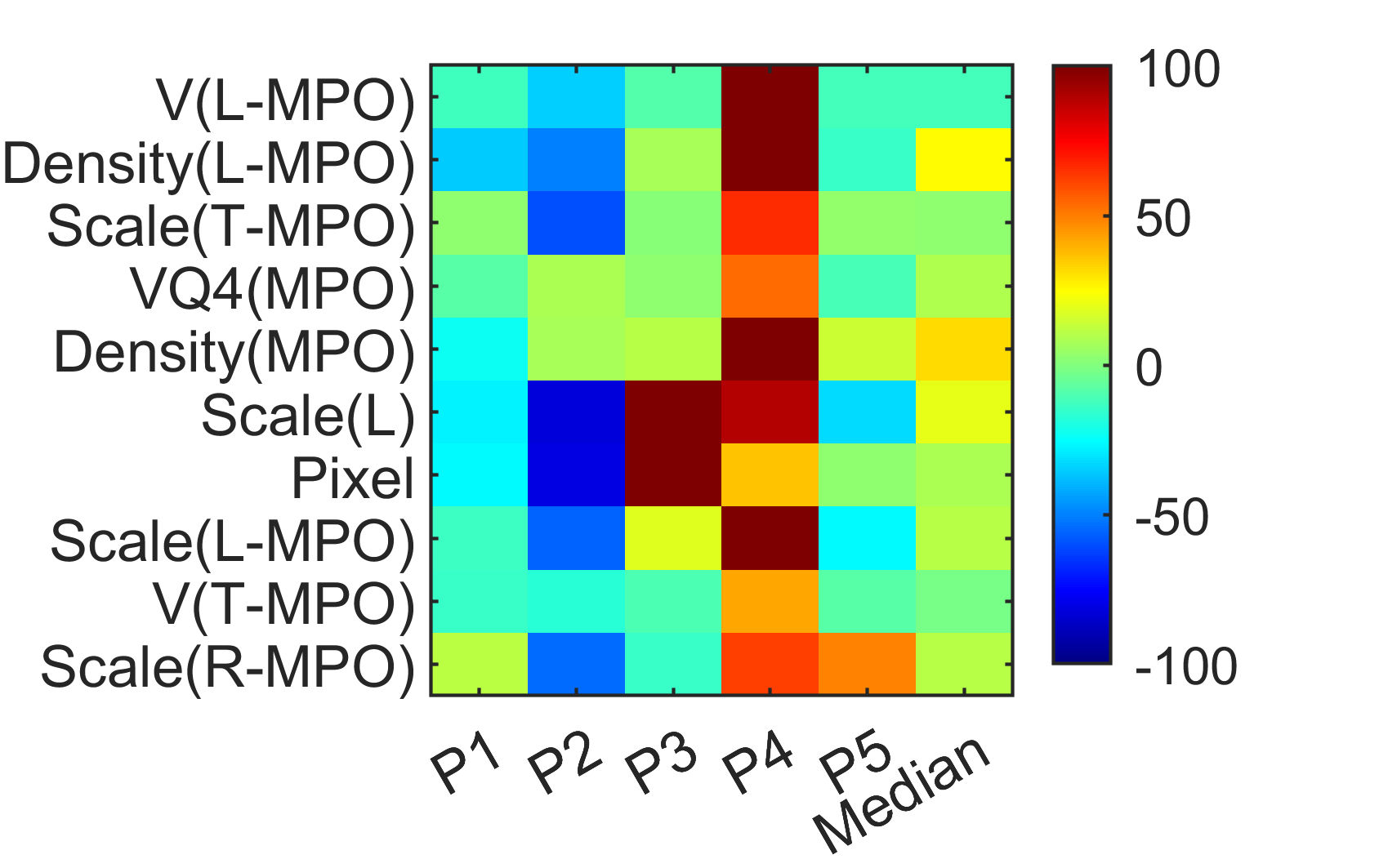}}{Read}
    \caption{ Change from normal to weakness (\%) in the top 10 behavioral features ranked among all participants in the activities. 
    {The color of a cell shows whether the feature value increased or decreased. As there are no consistent color changes when viewed horizontally, it is clear that different participants respond differently to the exercise.}
   {Compared to `Read,' the `Watch' activity exhibits less variation and presents more similar trends among participants. This suggests it might be a more reliable indicator of health state change. However, even within the same activity, individual participant's feature trends can vary significantly.}
    }
    \label{fig:bestAct}
\end{figure}

For further illustration, Fig. \ref{fig:bestAct} displays the changes in the top features in two activities that all participants have performed. 
`Watch' exhibits less variation (and more similar trends) among participants than `Read', suggesting it might be a better generalizable indicator of health state. 
It's important to note that even within the same activity, the trends in feature changes can vary significantly among participants.

Hence, it becomes evident that an individualized model tailored to each person would be more appropriate for effectively monitoring health states. Interpersonal differences consistently play a significant role in shaping our understanding of the true effects of a health condition \cite{s6}. Simulating conditions on healthy subjects and comparing within the same person allows for the removal of the influence of concurrent medical conditions and interpersonal differences, as different individuals possess distinct health baselines (e.g., gender, age, fitness level), and experience varying rates of condition progression. This approach helps to isolate and focus on the effects of weakness on individuals.

Several limitations are associated with this study. First, the exploration of environmental features remains incomplete due to constraints of the experimental design. The strong ``Time-of-Day'' features observed for participants P2 and P3 (see Fig. \ref{fig:fheach}) likely reflect their tendency to exhibit post-exercise weakness in the afternoon, as their workouts were typically completed in the morning. We plan to investigate such contextual biases in future work. In addition, factors such as room lighting, surrounding objects, and even weather conditions may have influenced the recordings, since all data were captured within participants' homes.

The recording setup may also have affected participants' naturalistic behavior \cite{angrosino2016naturalistic}. This effect is most visible in activity duration patterns (Table \ref{tab:record}), where most activities lasted less than 30 minutes when a camera was present. Such durations are shorter than what would typically occur in real-world settings.

The small sample size and imbalanced data distribution further limit population-level generalization. However, the longitudinal nature of the data supports robust personalized modeling, and no universal feature set was identified across participants.

Finally, although the simulation approach is effective for isolating behavioral phenotypes, it requires future validation using data collected from individuals with chronic weakness.

\section{Conclusion}

Physical weakness is a common manifestation of aging and chronic health conditions. It often appears as subtle changes in daily movement patterns. Detecting these behavioral shifts is difficult because many real-world factors can mask them. To address this challenge, we introduced a privacy-preserving sensing system that quantifies fine-grained behavioral changes linked to weakness. We used exercise-induced fatigue in healthy adults to simulate weakness. This controlled approach isolates the behavioral phenotype and provides personalized data for validating our system.

We monitored participants during common daily activities and quantified their shift from normal to post-exercise weakness. We developed detailed and semantically meaningful behavioral features. We identified which features and activities were most indicative of the weakness phenotype. We also examined temporal windows from 30 seconds to a full day. With optimal feature sets, activity contexts, and time scales, the Bayesian Network models specific to participants achieved an accuracy of 0.95 at an 8-hour interval and 0.97 at the daily level. Features such as altered movement on the non-dominant body side and inactivity distribution emerged as promising behavioral markers. Personalized models are recommended, as participants showed substantial behavioral variability. Participant-specific features outperformed unified feature sets, demonstrating feasibility for individualized anomaly detection.

This work shows that computer vision and machine learning can detect behavioral characteristics of simulated
weakness. The approach focuses on within-person dynamics, not population-level diagnosis. The simulation
provides controlled insight into patterns relevant to early functional decline. Although simulated weakness
differs from chronic conditions, it helps us study behavioral changes that may inform future clinical validation.
Ranking features and activities supports individualized monitoring. Time-scale analysis offers practical guidance
for deployment. The system protects privacy, and its interpretable features provide clearer insight than black box deep learning models.

To the best of our knowledge, this is one of the first systems to detect simulated physical weakness from visual behavioral cues while offering interpretable explanations. While our findings may not generalize across individuals, this limitation is consistent with the intended scope of personalized monitoring. Future work will integrate environmental context more deeply, and may involve longer-term, larger-scale studies with data from older adults for potential behavior signatures related to chronic health conditions, as well as including multi-camera or multi-room deployments.

\noindent\textbf{Acknowledgments.}
This research was funded by the Legal \& General Group (research grant to establish the independent Advanced Care Research Centre at the University of Edinburgh). The funder had no role in the conduct of the study, interpretation, or the decision to submit for publication. The views expressed are those of the authors and not necessarily those of Legal \& General.
Approval for the experiments was granted by the School of Informatics Ethics Committee.

\bibliographystyle{unsrt}
\bibliography{sample-base}

\end{document}